\documentclass[journal]{IEEEtran}
\usepackage{caption}
\usepackage{times}
\usepackage{epsfig}
\usepackage{cite}
\usepackage{graphicx}
 \usepackage{epstopdf}
\usepackage{amsmath}
\usepackage{amssymb}
\usepackage[marginal]{footmisc}
\usepackage{threeparttable}
\usepackage{stfloats}
\usepackage[pagebackref=true,breaklinks=true,letterpaper=true,colorlinks,bookmarks=false]{hyperref}
\usepackage{multirow}
\usepackage{bm}

\begin{document}

\title{Contrastive Proposal Extension with LSTM Network for Weakly Supervised Object Detection}

\author{Pei~Lv,
	Suqi~Hu, and 
	Tianran~Hao
	\thanks{\indent Pei Lv and Tianran Hao are with the School of Computer and Artificial Intelligence, Zhengzhou University, Zhengzhou 450001, China. E-mail: ielvpei@zzu.edu.cn; iehaotianran@gmail.com.}
	\thanks{\indent Suqi is with Henan Institute of Advanced Technology, Zhengzhou University, Zhengzhou 450003, China. E-mail: qusuhu@gs.zzu.edu.cn.}
}

\markboth{Journal of \LaTeX\ Class Files,~Vol.~14, No.~8, August~2015}%
{Shell \MakeLowercase{\textit{et al.}}: Bare Demo of IEEEtran.cls for IEEE Journals}

\maketitle

\begin{abstract}
   Weakly supervised object detection (WSOD) has attracted more and more attention since it only uses image-level labels and can save huge annotation costs. Most of the WSOD methods use Multiple Instance Learning (MIL) as their basic framework, which regard it as an instance classification problem. However, these methods based on MIL tend to only converge on the most discriminative regions of different instances, rather than their corresponding complete regions, that is, insufficient integrity. Inspired by the human habit of observing things, we propose a new method by comparing the initial proposals and the extended ones to optimize those initial proposals. Specifically, we propose one new strategy for WSOD by involving contrastive proposal extension (CPE), which consists of multiple directional contrastive proposal extensions (D-CPEs), and each D-CPE contains LSTM-based encoders and dual-stream decoders. Firstly, the boundary of initial proposals in MIL is extended to different positions according to well-designed sequential order. Then, the CPE compares the extended proposal and the initial one by extracting the feature semantics of them using the encoders, and calculates the integrity of the initial proposal to optimize its score. These contrastive contextual semantics will guide the basic WSOD to suppress bad proposals and improve the scores of good ones. In addition, a simple dual-stream network is designed as the decoder to constrain the temporal coding of LSTM and improve the performance of WSOD further. Experiments on PASCAL VOC 2007, VOC 2012 and MS-COCO datasets show that our method has achieved the state-of-the-art results.
\end{abstract}

\IEEEpeerreviewmaketitle

\section{Introduction}

\IEEEPARstart{w}{eakly} supervised object detection (WSOD), which only uses image-level labels to train the detection model, has attracted the wide attention of researchers. Compared with the fully supervised object detection (FSOD) task \cite{8721661,8434341,9305976,9502943}, which  requires bounding-box annotations for massive amounts of data, WSOD is relatively time-saving, labor-saving and promising, while still has some challenges to obtain satisfactory performance. 

Nowadays, most of the WSOD methods use Multiple Instance Learning (MIL) as their basic framework. They regard WSOD as an instance classification problem, and use the instance classifier to approach the goal of object detection. 
One of the most representative work is the Weakly Supervised Deep Detection Network (WSDDN) \cite{2016Weakly}, which firstly combines MIL with WSOD. Due to its effectiveness and conciseness, plenty of following work are further carried out based on it. The work in \cite{tang2017multiple,tang2018pcl} proposed an online refinement module to solve the problem that WSDDN only locates the object region with the highest salience score, rather than the whole object. However, WSOD methods are still far from achieving the competitive results generated by FSOD methods.
\begin{figure}[t]
    \centering
    \includegraphics[width=0.50\textwidth]{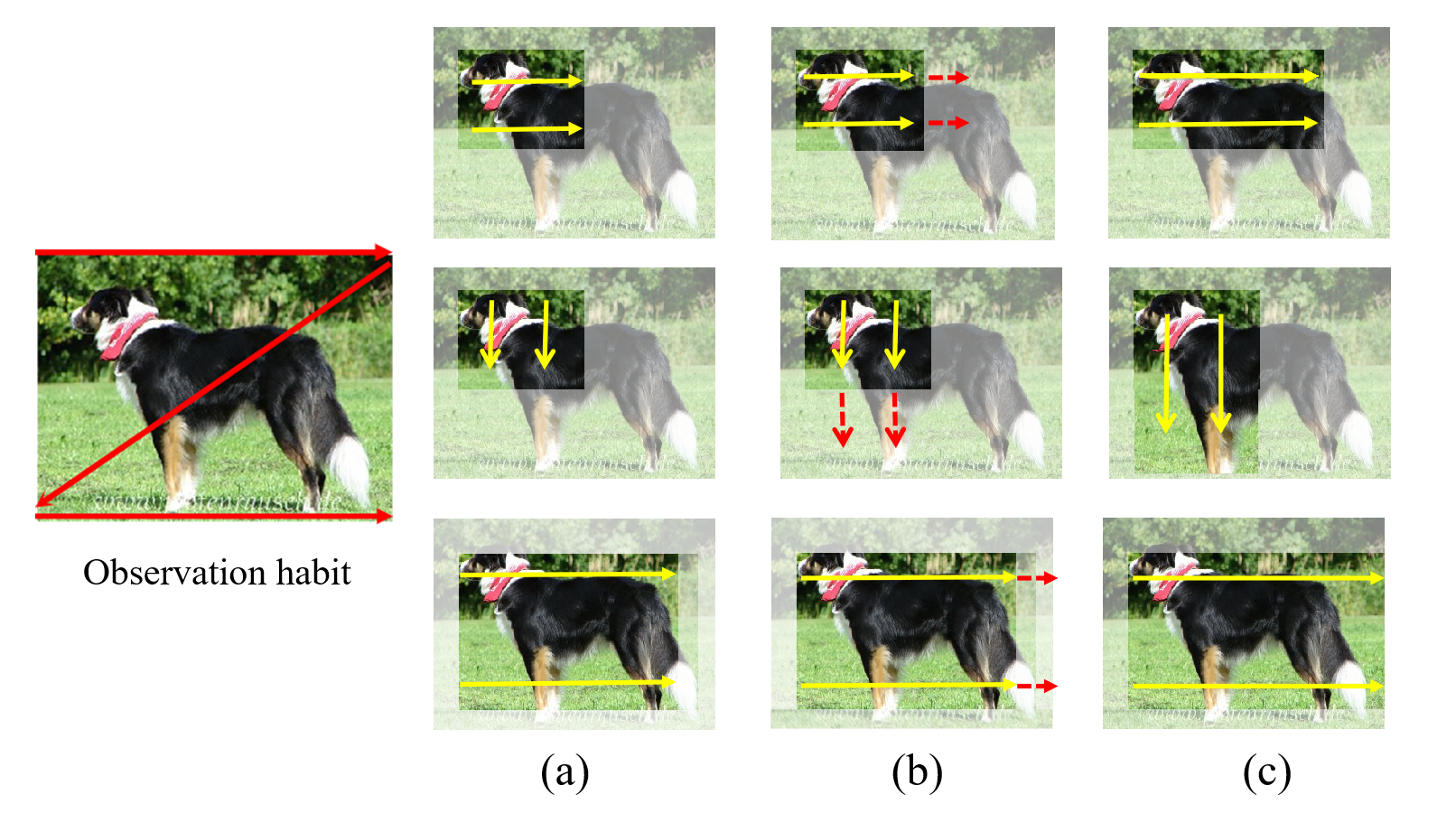}
    \caption{The scanning way in locating complete objects. (a) represents a certain part of the object in the observed scene, (b) represents the trend of subsequent observation based on previous column, and (c) represents the results of further observation. The instance in each row contains a certain part of the object, and its extensions in different directions.}

    \label{ff1}
    \end{figure} 

In the real world, we often take a scanning way to observe objects as shown in Fig. \ref{ff1}. When only the local part of the object is perceived, we will be in the habit of continuing the observation for further exploration and searching its remaining parts in a certain order. It is very suitable to model this sequential process using recurrent neural networks, such as LSTM network. In MIL, an image contains multiple different initial proposal regions, however, the objects in most of the proposal regions are incomplete. Therefore, when a single proposal is observed, we can regard it as a certain part of one object and make further observations of its neighboring area. Under these situations, this single instance can be involved to form a direct comparison with other further observed instances. The direct comparison will allow us to conduct an evaluation of the initial instance in two cases: 1) if the further observed region (extended proposal) is similar to the previous one (original proposal), the object in the observed region is likely to be incomplete. 2) if the further observed region is obviously different from the original one, the original proposal prefers to containing a complete object.

Inspired by the above facts, we propose a new Contrastive Proposal Extension (CPE) module for basic WSDDN framework, which mimics the scanning way of human beings to extract the contrastive contextual semantics of different instances. In CPE, each adjacent row or adjacent column in the proposal feature is regarded as successive time series data. In order to extract the semantic feature between them, we choose the Long Short-Term Memory (LSTM) network \cite{hochreiter1997long} as the basic component of CPE. The CPE module contains multiple Directional Contrastive Proposal Extension (D-CPE) sub modules, and each D-CPE is composed of two encoders based on LSTM network and two decoders based on dual-stream structure network. 

In detail, the initial proposal is expanded in different directions to produce multiple extended proposals, which are used to compare with the original proposal for the integrity evaluation of the original proposal. D-CPEs extract different RoI pooling features for the extended proposal and the initial one, and then encode them individually. In the meantime, the decoder is involved to constrain the hidden space of the LSTM encoder. The D-CPE further calculates the contrastive contextual semantics between the initial proposal and the extended one, where the larger the contextual semantic difference between them, the greater the probability that the initial proposal is relatively complete. Afterwards, a number of contrastive contextual semantics in different directions are combined to jointly estimate the probability of the completeness of this proposal. In the forward propagation process of the WSDDN network, it provides basic detection vectors and classification vectors, and the CPE module provides contrastive contextual semantics. The contrastive contextual semantics will guide the detection vectors and classification vectors in WSDDN to generate more reasonable confidence scores, that is, increasing the confidence scores of instances with high probability of completeness, and suppressing the confidence scores of instances with low probability. We report the experimental results on PASCAL VOC 2007, 2012 and MS-COCO datasets, and compare our method with other representative approaches. 

The main contributions are summarized as follows:
\begin{itemize}

\item We propose one new Contrastive Proposal Extension (CPE) module in WSOD, which estimates the contextual semantic difference by exploring the comparable relationship between the original instance and its corresponding extended ones.

\item We introduce the LSTM network to encode the contextual semantics into the framework of WSOD, and effectively alleviate the mismatching problem between instance score and instance integrity.

\item The experiments on PASCAL VOC 2007, PASCAL VOC 2012 and MS-COCO show that our proposed method has achieved the state-of-the-art results.

\end{itemize}

\section{Related work}
Most of the existing work regards the task of WSOD as one MIL problem, and trains the model with image-level labels. To further improve the performance, some methods try to combine weakly supervised semantic segmentation with WSOD \cite{shen2019cyclic}. Moreover, in some FSOD methods, Recurrent Neural Network also played a considerable role\cite{li2016attentive}. Therefore, in this section, we mainly review above representative related work.

\subsection{Weakly Supervised Object Detection}
In order to solve the task of only using image-level labels to realize object detection, most of existing work are proposed based on MIL framework. For such kinds of methods, an image can be regarded as an instance package containing multiple instances, and these instances will be used for model training to select the most reasonable candidate instances. These WSOD methods mainly consist of two consecutive stages, object discovery and instance refinement. The object discovery stage usually combines MIL and CNNs to implicitly model the localization of potential objects using image-level labels\cite{bilen2015weakly,ge2018fewer,gokberk2014multi,shi2016weakly,wang2014weakly,wang2015relaxed}. In addition, contextual information \cite{kantorov2016contextlocnet}, attention mechanism \cite{teh2016attention}, gradient map \cite{shen2019category} and semantic segmentation \cite{li2016weakly} are also used to mine the complete object proposal. To optimize the overall framework, some approaches tried to use minimum entropy prior \cite{wan2018min}, multi-view learning \cite{zhang2018ml}, and generative adversarial learning \cite{shen2018generative}. Beyond that, to alleviate the problem of non-convexity, Wan et al. \cite{wan2019c} introduced the continuous optimization algorithm into MIL, and proposed an effective continuation MIL (C-MIL) method. Arun et al. \cite{arun2019dissimilarity} designed the WSOD network based on incoherence coefficients by exploiting the difference between the distribution of unknown predictions and the label-aware conditional distribution. Chen et al. \cite{chen2020slv} proposed a spatial likelihood voting (SLV) to aggregate the process of proposal localizing, using voting results to regularize the bounding-box. To detect the complete region of objects, recent work began to focus on adopting cascaded refinements of MIL classifications \cite{tang2018pcl}, leveraging continuity strategies \cite{wan2018min} and modeling the location uncertainty\cite{arun2019dissimilarity}.

Contrastive learning \cite{9723577,9577781,2020Contrastive}, as a self-supervised learning method, in the absence of labels, can learn more general features from unlabeled data before performing tasks such as classification. After that, using other techniques to fine-tune the pre-trained model will effectively improve the object detection performance even better than supervised training and weakly supervised training with image-level labels.

Inspired by the idea of comparing the similarities and differences of data in contrastive learning, we use CPE to extract the contrastive contextual semantics between the initial proposal and the extended one, which is further combined with the semantic information of the backbone network to estimate a more reasonable score of the initial proposal.

\subsection{Weakly Supervised Semantic Segmentation }
Weakly supervised semantic segmentation (WSSS) task \cite{9465740,9440699} is closely related to the WSOD, and they are often combined together to achieve better performance for their respective goals. To solve the problem of the lack of object border annotations in WSOD, some recent work combined the segmentation map or class activation map with the WSDDN to obtain more complete objects. Shen et al. \cite{shen2019cyclic} proposed a joint weakly-supervised detection and segmentation framework into a multi-task learning framework. Diba et al. \cite{diba2017weakly} generated candidate regions based on the segmentation results, and used these regions for multi-instance learning. Similarly, Gao et al. \cite{gao2018c} used Affinity Net in \cite{ahn2018learning} to generate segmentation maps as an additional basis for object detection. Cheng et al. \cite{cheng2020high} used the segmentation maps generated by Grad-CAM \cite{selvaraju2017grad} to filter the initial proposal boxes, which reduced the interference of additional boxes and obtained initial regions with higher quality. In addition, Zeng et al. \cite{zeng2019wsod2} proposed a bottom-up and top-down object distillation framework (WSOD${^2}$) that fully combined with deep objectness representation learned from bottom-up evidences. Although the introduction of weakly-supervised semantic segmentation maps is conducive to detecting more complete objects, this kind of methods still can not extract all the contents of the image completely, and have certain limitations.

To avoid the problem caused by incomplete segmentation map and class activation map, we directly compare the features of original instances and the extended ones, which allows CPE to detect subtle differences between the extended and the initial region, further determining the integrity of the initial instance.

\begin{figure*}[t]
    \centering
    \includegraphics[width=0.95\textwidth]{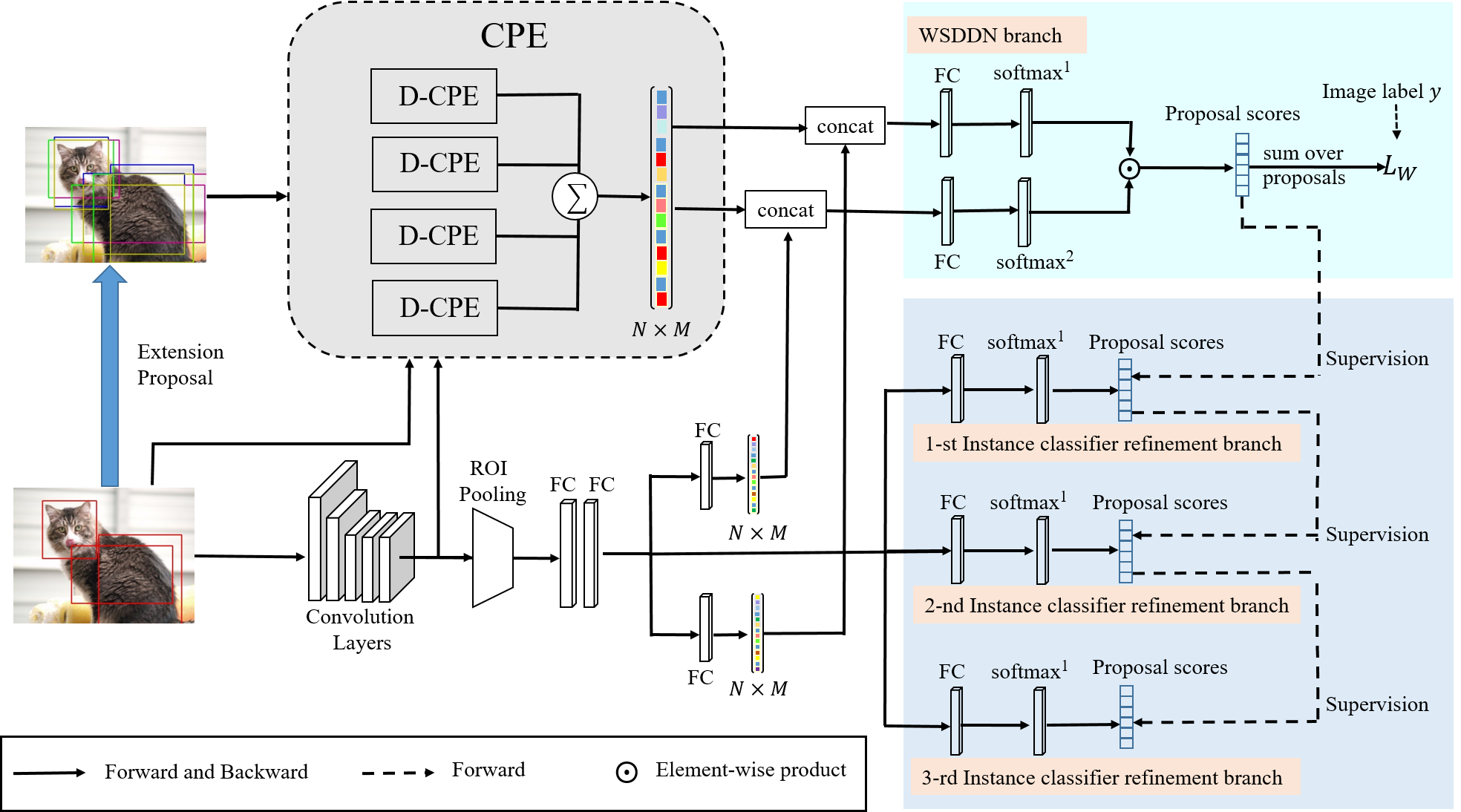}
    \caption{Overview of our weakly supervised object detection framework. A backbone network is used to extract image features, which are taken into the CPE module(Sec. \ref{3.2} and \ref{3.3}) and a separate RoI Pooling layer. The CPE uses image features to extract contextual semantics between the initial proposal and the extended one. This contrastive context is fused with the feature vector generated by the basic MIL module to jointly generate the proposal score, and the average classification score of the refinement branches is used to locate the object. $softmax^1$ is the softmax operation over classes and $softmax^2$ is the softmax operation over proposals.}
    \label{fig_overall}
\end{figure*} 

\subsection{Recurrent Neural Network}
Recurrent Neural Network (RNN) has a variety of forms, such as Bidirectional RNNs, LSTM, GRU, Graph LSTM, etc. They have a wide range of applications in different fields, including speech recognition, semantic segmentation, trajectory prediction, natural language processing and object detection. Yan et al. \cite{yan2016combining} adopted LSTM to construct a spatial loop layer to directly capture the global context and improve the feature representation for semantic segmentation. Bell et al. \cite{2016Inside} used RNN to construct a spatial recurrent neural network to integrate the contextual information outside the region of interest for object detection. In \cite{stewart2016end}, Stewart et al. used the LSTM network to decode the extracted features, generated time sequences, and performed end-to-end training to detect humans in dense areas. Li et al. \cite{li2016attentive} combined CNN with LSTM to propose an object detection model based on a novel attention to context convolution neural network (AC-CNN), which fused the global context with the local context to make the final detection. Liang et al. \cite{liang2016semantic} proposed a Graph Long-short Memory Network for semantic object parsing. Van et al. \cite{van2016pixel} proposed a Pixel Recurrent Neural Network to model the discrete probability of the pixel values and encode the complete set of dependencies in the image.

Inspired by the human habit of taking a scanning way to observe things in sequential order, we choose the relatively powerful and flexible LSTM as the basic structure of the CPE module among these various RNNs. In practice, although GRU is faster and simpler than LSTM, it always uses cell state indiscriminately, which is not conducive to extracting the contextual semantics between instances and their extensions.

\section{Proposed Approach}
Our work focuses on increasing the degree of matching of the initial proposal evaluated by the detector to improve the performance of WSOD. We build our method with online instance classifier refinement branches (OICR) as the basic framework, and the overall framework is shown in  Fig. \ref{fig_overall}. The CPE consists of multiple D-CPEs, which is responsible for extracting the contrastive semantics between the initial proposals and the extended ones in different directions. In the training stage, different D-CPEs will be fused together and combined with the semantic information of the basic framework to jointly estimate the final score of initial proposals.  

\subsection{Basic MIL Network}
\label{3.1}
Bilen et al. \cite{2016Weakly} firstly proposed a WSDDN framework for WSOD by combining with MIL. Given an image $I$, and corresponding image-level labels $y = [y_1,y_2,...,y_c]^T$, we can obtain initial proposals $B = \{B_1,B_2,...,B_N\}$ by WSDDN, where $C$ is the total number of categories, $y_c = 1$ or $0$ represents whether the image contains category $c$, and $N$ is the total number of initial proposals generated by proposal generation methods. Firstly, $I$ passes through the CNN backbone layer and RoI pooling layer to generate a series of proposal features for the initial proposals. And then, the proposal features are taken into two fully connected layers respectively to generate feature matrices $x^{cls}$, $x^{dec}$ $\in$ $\mathbb{R}^{N \times C}$. We apply $softmax$ functions for categories and proposals respectively:
\begin{equation}\begin{split}{[{\sigma}_{cls}(x^{cls})]}_{ij}=\frac{x_{ij}^{cls}}{{\sum\limits}_{k=1}^Cx_{ik}^{cls}},
\\
{[{\sigma}_{dec}(x^{dec})]}_{ij}=\frac{x_{ij}^{dec}}{{\sum\limits}_{k=1}^Nx_{kj}^{dec}},\end{split}\end{equation}where ${[{\sigma}_{cls}(x^{cls})]}_{ij}$ represents the ${j^{th}}$ class label for the ${i^{th}}$ region, and ${[{\sigma}_{dec}(x^{dec})]}_{ij}$ represents the learned weight for the ${j^{th}}$ class of the ${i^{th}}$ region. Afterwards, the basic MIL network generates proposal scores for all candidate regions, $x^s = {\sigma}_{cls}(x^{cls})\odot {\sigma}_{dec}(x^{dec})$.

In order to integrate the CPE module (referring to Sec. \ref{3.2} and \ref{3.3}) into the overall framework, we make some changes to the basic structure of MIL. In detail, the CPE module generates the contrastive contextual semantics $\mathcal{N}$ ($\mathcal{N}$ $\in$ $\mathbb{R}^{N\times C}$). Then, we merge $\mathcal{N}$ with $x^{cls}$ and $x^{dec}$ to generate new classification vectors and detection vectors, whose dimensions are consistent with that of $x^{cls}$ and $x^{dec}$. The formula is as follows:
\begin{equation}\begin{aligned}
x^{rcls}&=Linear([ x^{cls},\mathcal{N} ]), x^{rcls}\in \mathbb{R}^{N\times M},\\
x^{rdec}&=Linear([ x^{dec},\mathcal{N} ]), x^{rdec}\in \mathbb{R}^{N\times M}.\end{aligned}\end{equation}$Linear$ represents a single linear layer, which is to reduce the feature loss caused by the fusion of $\mathcal{N}$ and $x^{cls}$, $x^{dec}$. We use the similar calculation method as MIL, and via $x^{rcls}$, $x^{rdec}$, generate scores $x^s = \sigma_{dec}(x^{rdec})\odot \sigma_{cls}(x^{rcls})$ for all proposals. Finally, by summing the scores of all instances for category $c$, we can obtain the scores of category $c$ in the image: $\sigma^c = \Sigma_{i=1}^Nx_{ic}^s$, and we optimize the multi-class across entropy loss in Eq. \ref{eq_loss} to train the basic WSDDN.
\begin{equation}\label{eq_loss}L_W=-\sum\limits_{c=1}^C y_c\log\sigma^c + (1-y_c)\log(1-\sigma^c).\end{equation}

Furthermore, we introduce multi-layer online instance classifier refinement branches in \cite{tang2017multiple} to obtain a more reasonable instance classifier. The output of refinement branches $\varphi^k \in \mathbb{R}^{N \times (C+1)}$ are different from the basic instance classifier, where $(C+1)$ represents $C$ object classes and an additional background class, $k$ represents the $k^{th}$ instance classification optimizer. Based on the output $\sigma$ of the previous basic MIL instance classifier, a more reasonable instance score can be obtained by using the online classifier branch. Assuming that the image contains class $c$, the $k^{th}$ instance classification optimizer will select the instance $B_c^k$ with highest score in the proposal region of class $c$, and then calculates the IoU formed by surrounding areas $B$ and $B_c^k$. When the filter is larger than the threshold $\tau$, the region is marked as class $c$, otherwise, it will be marked as background class. The loss function of a single instance classifier is:

\begin{equation}L_r^k = -\frac{1}{|{N}|}\sum\limits_{r=1}^{N}\sum\limits_{c=1}^{C+1} y_{cr}^k\log{\varphi_{cr}^k},\end{equation}
where $y_{cr}^k$ represents the $c^{th}$ class label for the $r^{th}$ proposal in the $k^{th}$ instance classification optimizer, and ${\varphi_{cr}^k}$ represents the score for the $c^{th}$ class of the $r^{th}$ proposal in the $k^{th}$ instance classification optimizer.

\begin{figure*}[t]
    \centering
    \includegraphics[width=0.95\linewidth]{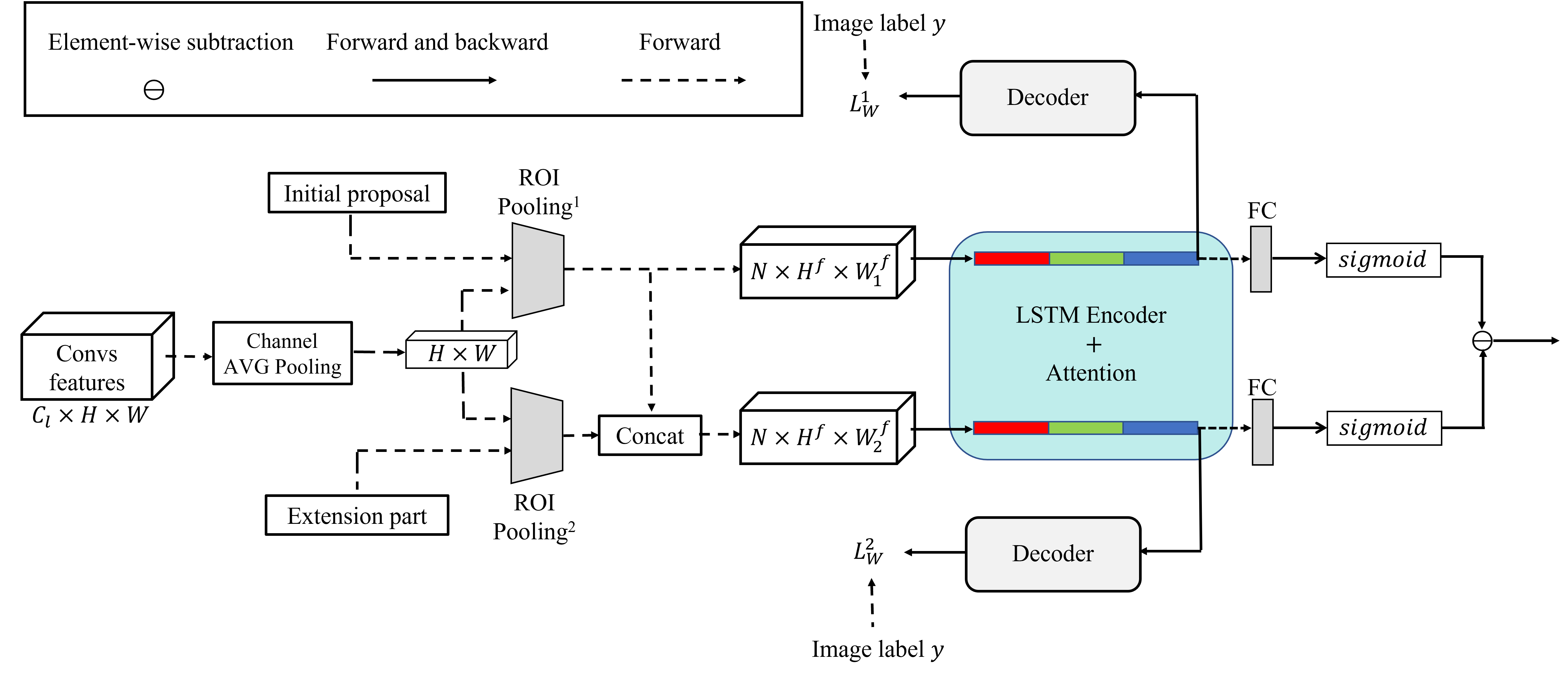}
    \caption{Illustration of D-CPE. The decoder of D-CPE is based on a small WSDDN dual-stream branch, and its detailed structure is in Sec.\ref{3.3}. $RoIPooling^1$ and $RoIPooling^2$ represent two pooling layers with different pooling sizes.}
    \label{D-CPE}
\end{figure*}

\subsection{Proposal Extension}
\label{3.2}
In our method, D-CPE module is in charge of extracting the contrastive contextual semantics between the extended instance and the initial one. As shown in Fig. \ref{D-CPE}, each D-CPE contains two RoI pooling layers, two LSTM encoders, and corresponding decoders. The LSTM encoder encodes proposal features to obtain corresponding semantic information, and the decoder is composed of two dual-stream branch networks to constrain the coding space of the LSTM encoder.  

First, the initial region $B_i(x_i,y_i,w_i,h_i)$ will be used to obtain the extended proposal, where $x_i$, $y_i$ are the initial left-top coordinates of the ${i^{th}}$ proposal region in the image $I(W^{I},H^{I})$. $w_i$, $h_i$ are the width and height of the proposal region, $W^{I}$, $H^{I}$ are the width and height of the image. We respectively expand the boundaries in the four directions $D={\{R2L, L2R, T2B, B2T\}}$ of the initial region to form extended regions $B_L$, $B_R$, $B_B$, $B_T$. $R$, $L$, $T$ and $B$ are the abbreviations of right, left, top and bottom directions respectively. $B_L$ represents the extended region formed by keeping the right border of the initial region unchanged and extending the left border of the initial region along the left direction. $B_L$, $B_R$, $B_T$, $B_B$ have similar meanings, but extend in different directions, and eventually formulate the following:\begin{equation}\begin{aligned}
&B_L(max(x_i - w'_i ,0),y_i,w_i+min(w'_i,x_i) ,h_i),\\
&B_R(x_i,y_i,min(W^{I}-x_i+1,w'_i+w_i),h_i),\\
&B_T(x_i,max(y_i-h'_i,0),w_i,h_i+min(h'_i,y_i)),\\
&B_B(x_i,y_i,w_i,min(H^{I}-y_i+1,h'_i+h_i)),
\label{5}
\end{aligned}\end{equation}where $w'_i = \frac{(w_i)^2}{(h_i*t)}$, $h'_i = \frac{(h_i)^2}{(w_i*t)}$, $t$ is a hyper-parameter and $t > 1$. They can be abbreviated as $B_L = (B:B_l)$, $B_R = (B:B_r)$, $B_T = (B:B_t)$, $B_B = (B:B_b)$, where $B_l$, $B_t$, $B_r$, $B_b$ are the extra areas after the extension except the initial region. In the entire training process, the extension strategy will only be performed once on the initial proposal to obtain the extended ones. $x_i$, $y_i$ represent the left-top coordinates of the proposal $i$. $w_i$, $h_i$ represent the width and height of this proposal. $w'_i$, $h'_i$ are the extended width and height according to the aspect ratio of the proposal $i$. $W^{I}$, $H^{I}$ represent the width and height of the image. Taking the extension in R2L direction as an example, when the left boundary of the proposal exceeds the image, that is, when $(x_i - w'_i) < 0$, and the actual extension width is $x_i$, the left boundary of the image is used as the extended border, otherwise, $x_i-w'_i$ is used as its left border, and the actual extension width is $w'_i$. The extended borders in other directions are calculated in a similar way. 

\subsection{Contrastive Contextual Estimation}
\label{3.3}
Extended proposals in different directions are taken into different D-CPE respectively with initial proposals. Then, these proposals and the convolutional features of the image are taken into the RoI pooling layers in D-CPE to be further processed. The outputs of the RoI pooling layer are respectively sent to the corresponding LSTM encoders to obtain semantic information. 

To explore along a certain direction gradually, each adjacent column or row in the pooling feature is regarded as the corresponding input according to a certain order, and is used to extract the semantic differences to judge the integrity of the initial region. Inspired by other work encoding image features using RNN \cite{2019Seeing}, \cite{2021Pix2seq}, \cite{2016Inside}, we choose the Long Short-Term Memory (LSTM) network \cite{hochreiter1997long} with the best performance in our experiments as the basic component of CPE.

The semantic information generated by different encoders will be fed into a linear layer with $sigmoid$ activation function to obtain the corresponding semantic scores. The contrastive semantic score will be merged with the features generated by the basic dual-branch structure network to adjust score of the initial proposal. Incomplete proposals will gain lower score, and relatively complete proposals will obtain higher score. The scores of all the proposals will be optimized by fusing the contrastive semantic score and the score estimated by the original WSDDN. The specific process is described as follows.

We extend the initial proposal in different directions to obtain additional extended ones. The features which are input to the CPE need to be processed additionally. We perform channel average pooling on the extracted features and generate the new features $X^F(m,n)=\frac{1}{C_l}\Sigma_1^{C_l}X'_{m,n}$, where $X'_{m,n}\in \mathbb{R}^{C_l\times H\times W}$ is the feature value at the position $(m, n)$ in the feature map extracted by the $i^{th}$ channel, $i \in \mathbb\{1,2,...,C_l\}$, and $C_l$ is the number of feature channels.

The four D-CPEs are CPE$_L$, CPE$_R$, CPE$_B$, CPE$_T$ respectively. The composition of these four modules are exactly the same, and we take CPE$_L$ as an example for illustration. Given the initial proposal region $B$ and the extended region $B_L$, the feature $X^F$ are used as the input of two RoI Pooling layers with different parameters for the initial region proposal and the extended one.

\begin{equation}\begin{aligned}
X^B&=RoIPooling^{1}(X^F;B),\\
X^{B_l}&={RoIPooling}^{2}(X^F;B_l),\end{aligned}\end{equation}
where $X^B\in \mathbb{R}^{N\times{H^f}\times{W^f_1}}$, $X^{B_l}\in \mathbb{R}^{N\times{H^f}\times{w}}$, $N$ is the number of proposals, $H^f$ is the height of the pooling features, and $W^f_1$, $W^f_2$ are the width of the pooling features, $W^f_2 = W^f_1+w$. $W^f_1$ represents the width of pooling features of the initial region, $w$ represents the width of pooling features of the extra extended region. Due to the inherent characteristics of RoI pooling, we can obtain the features of the extended proposal by concatenating ${X^B}$ and ${X^{B_l}}$:
\begin{equation}\begin{aligned}
X^{B_L}&=X^{B}:X^{B_l},
\end{aligned}\end{equation}
where $X^{B_L}\in \mathbb{R}^{N\times{H^f}\times{W^f_2}}$.

\begin{figure}[t]
    \centering
    \includegraphics[width=0.50\textwidth]{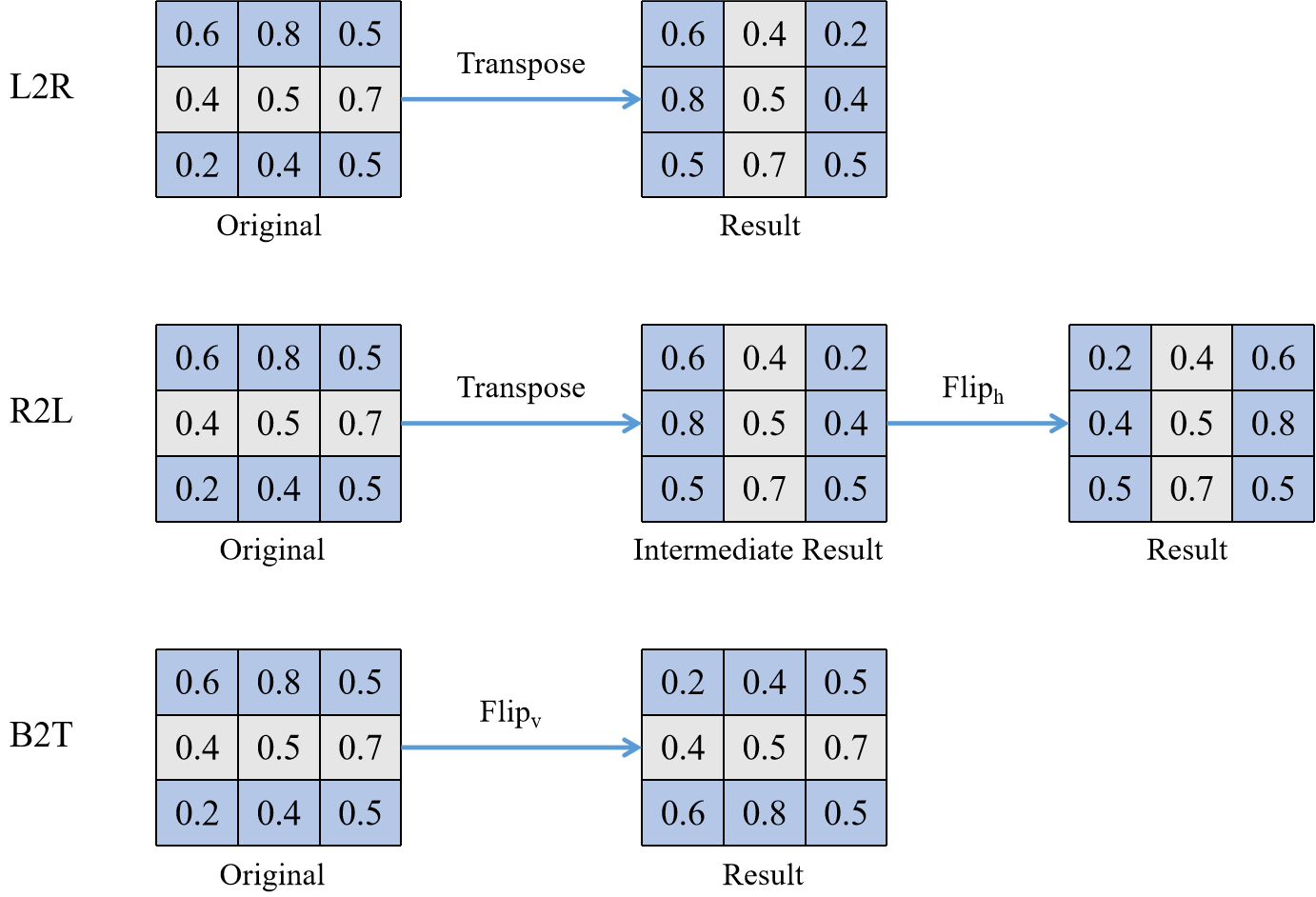}
    \caption{Transformation operations of pooling features extending in different directions. $Flip_h$ is the flip operation in the horizontal direction, and $Filp_v$ is the flip operation in the vertical direction. It does not need to perform transpose and flip operations for T2B.}
    \label{trans}
\end{figure}

The features of the extended proposal and the original one are respectively used as the input of the two encoders. The data in each row or column in the feature matrix will be regarded as the input of LSTM in specific direction $D$, $D={\{R2L, L2R, T2B, B2T\}}$. In order to input the feature matrix along the four directions into the LSTM encoder, some changes need to be done. As shown in Fig.\ref{trans}, L2R needs to perform a transpose operation, while R2L needs to perform a horizontal flip operation additionally.

After above operations, these vectors will be fed into LSTM encoders and perform sequential coding. For the CPE module, we still take the module CPE$_L$ as an example to demonstrate D-CPE, which is shown in Fig. \ref{D-CPE}. 
 $X_{r2l}^{B_i}\in \mathbb{R}^{N\times H^f\times 1}$ and ${X_{r2l}^{B_{Lj}}}\in \mathbb{R}^{N\times H^f\times 1}$ are respectively fed into the two LSTM encoders in CPE$_L$, where ${i=\{1,2,...,W^f_1\}}$, ${j=\{1,2,...,W^f_2\}}$, $W^f_1<W^f_2$. $X_{r2l}^B$, $X_{r2l}^{B_L}$ can be regarded as time sequences with $W^f_1$, $W^f_2$ steps, respectively.

\begin{equation}\begin{aligned}
D^B&=LSTM(X_{r2l}^B;W^f_1),\\
{D^{B_ L}}&=LSTM({X_{r2l}^{B_L}};W^f_2),\end{aligned}\end{equation}
where $W^f_1$, $W^f_2$ represent the different step sizes of the LSTM encoder, $W^f_2>W^f_1$. According to the characteristics of LSTM that the output of previous time step is used as the input of the next one, we further derive the vectors $D^B\in \mathbb{R}^{N\times M }$, ${D^{B_L}\in \mathbb{R}^{N\times M}}$, $M$ is a hyper-parameter, then we can infer:

\begin{equation}
\label{eq13}
\begin{aligned}
{D^{B_L}}&=LSTM({X_{r2l}^{B_L}};{W^f_2})\\
&=LSTM(X_{r2l}^B:{X_{r2l}^{B_l}};W^f_1+w),\\
\end{aligned}
\end{equation}
where $W^f_1$ represents the first $W^f_1$ column of $X_{r2l}^{B_L}$, and $w$ represents the last $w$ column, so $X_{r2l}^B$ can be regarded as the input of the previous $W^f_1$ step, and $X_{r2l}^{B_l}$ be regarded as the input of the next $w$ step.

\begin{figure}[t]
    \centering
    \includegraphics[width=0.50\textwidth]{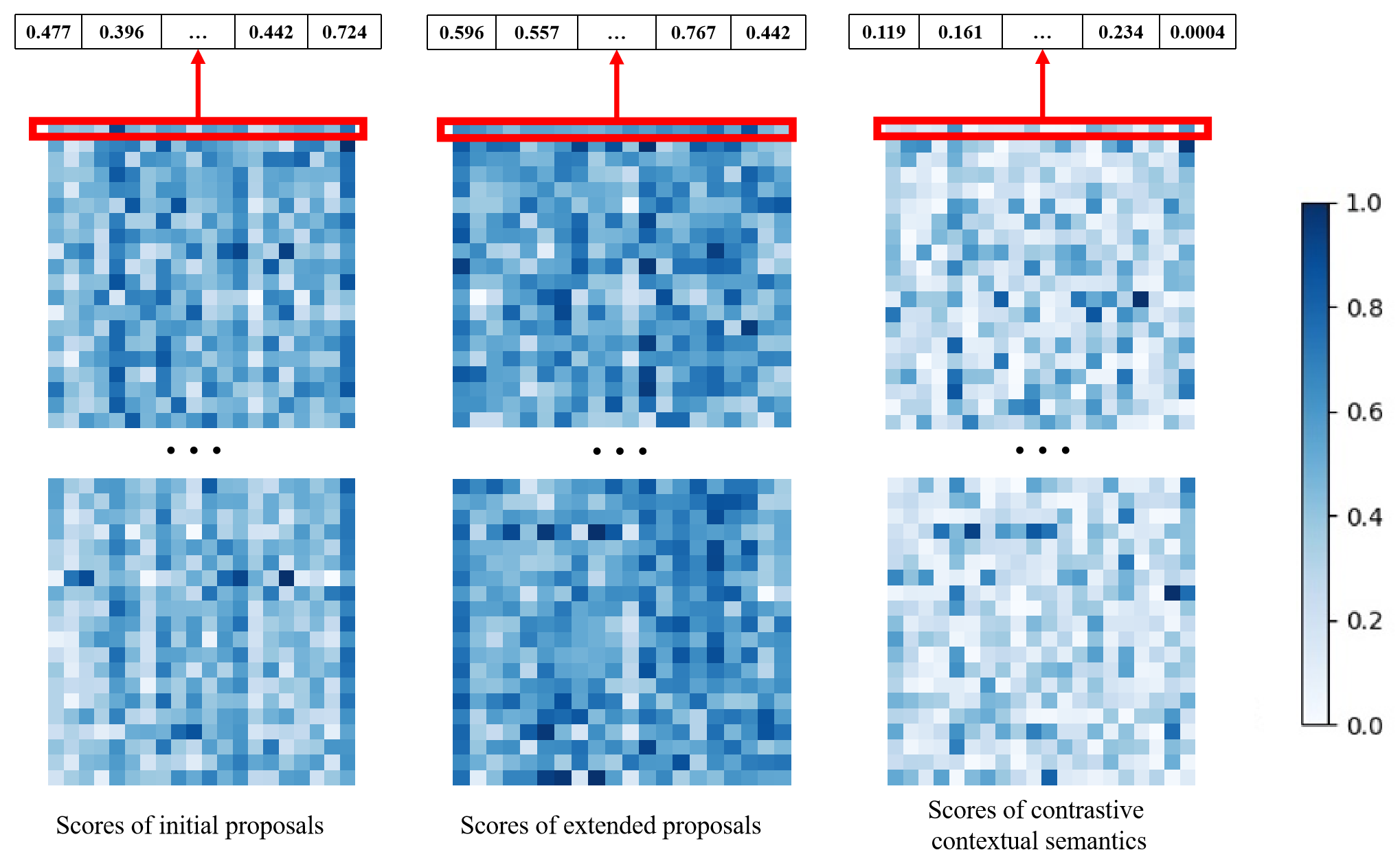}
    \caption{From left to right, the three columns represent the scores of the initial proposal, the extended proposal, and the contrastive contextual semantics obtained in D-CPE, respectively. Their dimensions are $N\times C$, $N$ represents the total number of proposals, and $C$ represents the number of categories.}
    \label{score_vis}
\end{figure}

 According to Eq. \ref{eq13}, we can infer that there is a relationship between $D^B$ and ${D^{B_L}}$ based on the additional extended region $B_l$. To further extract the difference, we use a separate linear layer and a $sigmoid$ layer to extract contextual semantics of proposals for each LSTM encoder. Moreover, we simply apply an attention mechanism \cite{vaswani2017attention} to better encode global information and extract complete semantic information. Then, we use the initial proposal contextual semantics score $S^B$ and the extended proposal contextual semantics score ${S^{B_L}}$ to construct $\mathcal{N}^L=|S^B-{S^{B_L}}|$ for each instance and the corresponding extended proposal. As shown in the Fig. \ref{score_vis}, the three columns from left to right represent the initial proposal score matrix, extended proposal score matrix and contrastive contextual semantic score matrix with the size of $N\times C$. Each row in the matrix represents the corresponding scores of each category. After obtaining the integrity scores of the initial proposal and the extended one, we perform element-wise subtraction operations on the two score matrices and absolute value calculations to obtain a preliminary contrastive semantic score. We further improve $\mathcal{N}^L$:
 \begin{equation}{\mathcal{N}^L} = \frac{{|{S^B} - {S^{B_L}}| - \min (|{S^B} - {S^{B_L}}|)}}{{\max (|{S^B} - {S^{B_L}}|) - \min (|{S^B} - {S^{B_L}}|)}}.\end{equation}

Subsequently, we combine the contrastive sores obtained by all D-CPEs by the following formula:
\begin{equation}{\mathcal{N}}={{\alpha}(\mathcal{N}^L}+{\mathcal{N}^R})+(1-{\alpha})({\mathcal{N}^B}+{\mathcal{N}^T)},\end{equation}
where $\alpha$ $\in$(0,1), ${\alpha}$ is a hyper-parameter. The semantic difference between the output of the two encoders in each D-CPE is mainly caused by the extension part ${B_l}$. When the semantic information of the initial proposal and extended one is quite different, the value of $\mathcal{N}^L$ will be large. ${B_l}$ may contain objects or backgrounds that are significantly different from ${B}$, and may not contain other components of the object in the initial proposal. That means ${B}$ already contains the complete object. On the contrary, when the semantic information of the initial proposal and extended one is similar, the value of $\mathcal{N}^L$ will be small, ${B_l}$ may be similar to ${B}$. The semantic information of the extended proposal is similar to the initial one, ${B_l}$ may contain other components of the object and the object in the initial proposal is incomplete.

To constrain the LSTM encoder, we add a small WSDDN dual-stream branch decoder to each LSTM encoder. Similar to the WSDDN network, the detection branch and the classification branch respectively include a linear layer and a $softmax$ layer in different directions. The loss function of CPE$_L$ during the training process is defined as follows:
\begin{equation}
    L_{CPE}(L) = L_W^1(L) + L_W^2(L) 
\end{equation}
$L_{W}^1, L_{W}^2$ are calculated in the same way as Eq.\ref{eq_loss}. Similarly, we perform the same calculation on other modules, the loss function of the complete CPE module is $L_{CPE} = (L_{CPE}(L) + L_{CPE}(R) +  L_{CPE}(B) + L_{CPE}(T))/4$. Meanwhile, in order to constrain the weakly supervised object detector, we integrate all the loss functions, and the final loss function of the entire model is as follows:
\begin{equation}L = {L_{CPE}} + {L_W} + \sum\limits_{k = 1}^3 {L_r^k}.\end{equation}

\section{Experiments}

In this section, we first describe the datasets, evaluation metrics, and the implementation details. Then, we conduct ablation studies to evaluate the influence of different modules or parameters on the results. Finally, we compare our proposed method with other state-of-the-arts on WSOD task to verify its effectiveness.

\subsection{Datasets and Evaluation Metrics}

We evaluate our method on the PASCAL VOC 2007 and 2012 \cite{2015The}, and the MS-COCO 2014 \cite{2014Microsoft} datasets. We only utilize image-level labels during model training. PASCAL VOC 2007 and 2012 datasets have 9,962 and 22,531 images. We choose the $trainval$ set (5,011 images for 2007 and 11,540 images for 2012) to train our model and choose $test$ set (4,951 images for 2007 and 10,991 images for 2012) to evaluate our model. For testing, we adopt two evaluation metrics, the mean of Average Precision (mAP) \cite{everingham2010pascal} on the $test$ set, and the correct localization (CorLoc) \cite{deselaers2012weakly} on  the $trainval$ set. Both these two metrics follow the standard PASCAL VOC protocol, IoU\textgreater0.5 between ground truth and predicted boxes. The MS-COCO dataset contains 80 categories, we choose the $train2014$ set (about 80K images) for training, $val2014$ set (about 40K images) and $minival$ set (5K images) for testing. We report AP@.50 and AP@[.50: .05: .95] on $val2014$ and $minival$.

\subsection{Implementation Details}

Our framework is based on VGG16 network \cite{simonyan2014very} pretrained on ImageNet. In detail, following the work in \cite{tang2018pcl}, we apply Selective Search (SS) \cite{uijlings2013selective} and MCG \cite{2016Multiscale} which generate about 2,000 proposals per image for the PASCAL VOC and MS-COCO datasets, respectively. We refine the instance classifier three times, i.e., $K=3$ and $\tau=0.1$ in Sec. \ref{3.1} according to \cite{tang2017multiple}. During training, the mini-batch size for SGD is set to 1, the momentum and weight decay are all set to 0.9, the learning rate is set to 5e-4 for the first $25k$, $40k$, $320k$ iterations and then decreased to 5e-5 in the following $35k$, $70k$, $480k$ iterations on VOC 2007, VOC 2012, MS-COCO respectively. We set the hyper-parameter $t=4$, ${\alpha}=0.5$. Specifically, we adopt five image scales \{480, 576, 688, 864, 1200\} and  horizontal flipping randomly on the images to generate 10 groups of augmented data. NMS $=0.3$ (with 30\% IoU threshold) is applied to compute the AP and CorLoc.

The experiments are implemented on the Pytorch deep learning framework, and part of the code is written in C++. All of our experiments are running on a workstation with an Intel(R) Xeon(R) CPU E5-2698 v4(2.20GHz) and 4 NVIDIA Tesla V100 GPUs.

\begin{table*}[t]
\renewcommand\arraystretch{1.2}
\caption{Results of D-CPE in different directions on PASCAL VOC 2007 $test$ set. “L2R”, “R2L”, “T2B” and “B2T” describe the extended proposal in different directions respectively. “ratio” means the aspect ratio of the proposal in Eq. \ref{5}.}
\label{ablation}
\scalebox{0.75}{
    \begin{tabular}{l|cccccccccccccccccccc|c}
    \hline
        \textbf{Directions} & \textbf{aero} & \textbf{bike} & \textbf{bird} & \textbf{boat} & \textbf{bottle} & \textbf{bus} & \textbf{car} & \textbf{cat} & \textbf{chair} & \textbf{cow} & \textbf{table} & \textbf{dog} & \textbf{horse} & \textbf{mbike} & \textbf{person} & \textbf{plant} & \textbf{sheep} & \textbf{sofa} & \textbf{train} & \textbf{tv} & \textbf{mAP} \\
    \hline
    \hline
       L2R 	&64.1	&77.6	&54.4	&26.5	&32.6	&74.0	&66.1	&68.2	&31.3	&60.6	&46.6	&61.0	&59.3	&69.3	&\textbf{19.8}	&30.0	&52.7	&60.5	&70.7	&56.4	&54.1\\
       R2L &63.9	&76.2	&55.6	&21.9	&33.4	&72.7	&62.6	&74.1	&32.3	&66.4	&46.4	&61.7	&62.0	&68.6	&15.4	&30.5	&53.3	&60.4	&\textbf{74.6}	&53.0	&54.3\\
       T2B &63.1	&77.2	&54.2	&32.3	&33.0	&72.3	&64.3	&70.4	&33.7	&62.2	&48.0	&62.4	&60.3	&70.0	&17.1	&31.6	&51.3	&60.1	&70.2	&51.2	&54.2\\
       B2T 	&60.0	&\textbf{78.0}	&55.3	&32.9	&\textbf{35.6}	&74.0	&\textbf{66.4}	&71.7	&32.1	&63.5	&\textbf{48.3}	&53.6	&62.1	&69.3	&19.4	&\textbf{32.6}	&52.2	&60.4	&72.2	&53.1	&54.6\\
       L2R+R2L &63.3	&77.8	&56.1	&30.3	&33.9	&71.3	&64.4	&\textbf{74.3}	&31.8	&63.8	&46.7	&63.3	&62.8	&69.7	&16.6	&31.5	&50.4	&\textbf{61.0}	&72.8	&57.0	&54.9\\
       T2B+B2T &62.9	&76.0	&58.4	&32.1	&35.0 	&\textbf{74.2}	&65.5	&73.1	&33.2	&63.9	&47.3	&61.3	&\textbf{63.0} 	&69.4	&13.4	&31.0	&52.6	&60.0	&71.4	&49.2	&54.6\\
       L2R+R2L+T2B+B2T &\textbf{66.1}	&75.8	&58.1	&32.0 	&28.8	&71.0	&65.9 	&71.4	&32.4	&64.8	&46.4	&65.5	&61.9	&70.4	&17.6 	&30.4	&53.4	&58.3	&72.6	&60.0	&55.1\\
       L2R+R2L+T2B+B2T+ratio &62.4	&76.4	&\textbf{59.7}	&\textbf{33.8}	&28.7	&71.7	&66.1	&72.2	&\textbf{33.9}	&\textbf{67.7}	&47.6	&\textbf{67.2}	&60.0	&\textbf{71.7}	&18.1	&29.9	&\textbf{53.8}	&58.9	&74.3	&\textbf{64.1}	&\textbf{55.9}\\
    \hline
    \end{tabular}%
     }
     \newline
   \label{gamma_h}
\end{table*}

\begin{table}[t]
\captionsetup{justification=centering}
\caption{Results with different CPE components. “Base” represents the case only with the LSTM encoder, “Attention” represents the case with self-attention mechanism, and “Decoder” represents the case with a simple dual-stream structure decoder for encoder.}
\centering
\scalebox{1.2}{
\renewcommand\arraystretch{1.2}
    \begin{tabular}{c|c|c|c}
    
    \hline
        \textbf & \textbf{Attention} & \textbf{Decoder}   & \textbf{mAP} \\
    \hline
    \multirow{4}*{Base} &     &     &50.4\\
          &      \checkmark      &                 &50.0 \\
    &     &       \checkmark           &51.2  \\
          &       \checkmark       &\checkmark            &55.9  \\
    \hline
    \end{tabular}%
    }
    \newline
\label{ablation_A}
\end{table}

\subsection{Ablation Studies }

We conduct some ablation studies to validate the effectiveness of our method, including the influence of  different directional extensions of CPE, various components of CPE, different decoders of D-CPE, the coupling coefficient of D-CPE, and extension coefficient of the extended proposal.

{\bf Different directional extensions.} We conduct ablation studies on different directional extensions in CPE module. As shown in Tab. \ref{ablation}, in row 1, 2, 3, 4, we discuss the effects in different directional extensions. Among these four extensions, although the result on one single direction is similar, the performance in different categories is different. The APs of $person$ category on L2R, R2L, T2B, B2T are 19.8\%, 15.4\%, 17.1\%, 19.4\%, respectively. The $person$ category is usually manifested in a standing posture in the scene, leading to the extension in the vertical direction, which is better than that in the horizontal direction. Meanwhile, there is a large difference in the extension of $dog$ category in two vertical directions. The reason is, compared with other animals, the limbs in the running and standing postures of $dog$ occupy most of the object region. In the extension along the B2T direction, the proportion of the object such as $aeroplane$, $chair$, $dog$, $motorbike$, is much smaller than that of the background, resulting in the extension along the T2B direction far better than that of B2T. 

Moreover, taking the combination of T2B and B2T as an example, Tab. \ref{ablation} shows that the overall result of T2B+B2T is not higher than that of B2T, but nearly equal. After analyzing 20 categories, we find that the results of $bike$, $boat$, $bottle$, $car$, $table$, $person$, $plant$, $sofa$, $train$, $tv$ are not higher than that of B2T after combining the expansion in the opposite direction T2B+B2T. The reason is that more discriminative features of these categories are distributed in the lower part of the overall object area. When the proposal expands from the bottom to top direction (B2T), the extended proposal has covered most of the object, and its lower boundary is close to the object boundary. After combining the extensions in the opposite direction (T2B), the lower boundary of the extended proposal may go beyond the object boundary and cover areas that do not belong to the object, causing to generate a lower score. Therefore, the overall result of T2B+B2T is a little lower. To be noted that, in some initial proposals, the usage of the extension in the opposite direction will conflict with the integrity evaluation of the initial proposal and cause the detector to make an unreasonable evaluation. After introducing more extension directions, this problem will be alleviated to a certain extent. In general, different directional extensions of CPE module will produce various results. The experiments also show that the more extension direction involved, the more complete comparative semantics extracted by the CPE, and the evaluation of the initial instance will be more accurate. Moreover, by introducing the aspect ratio of the proposal to adjust different scales of the extension operation dynamically, we can obtain more accurate extended instances to have better results.

\begin{figure*}[t]
    \centering
    \includegraphics[width=0.95\textwidth]{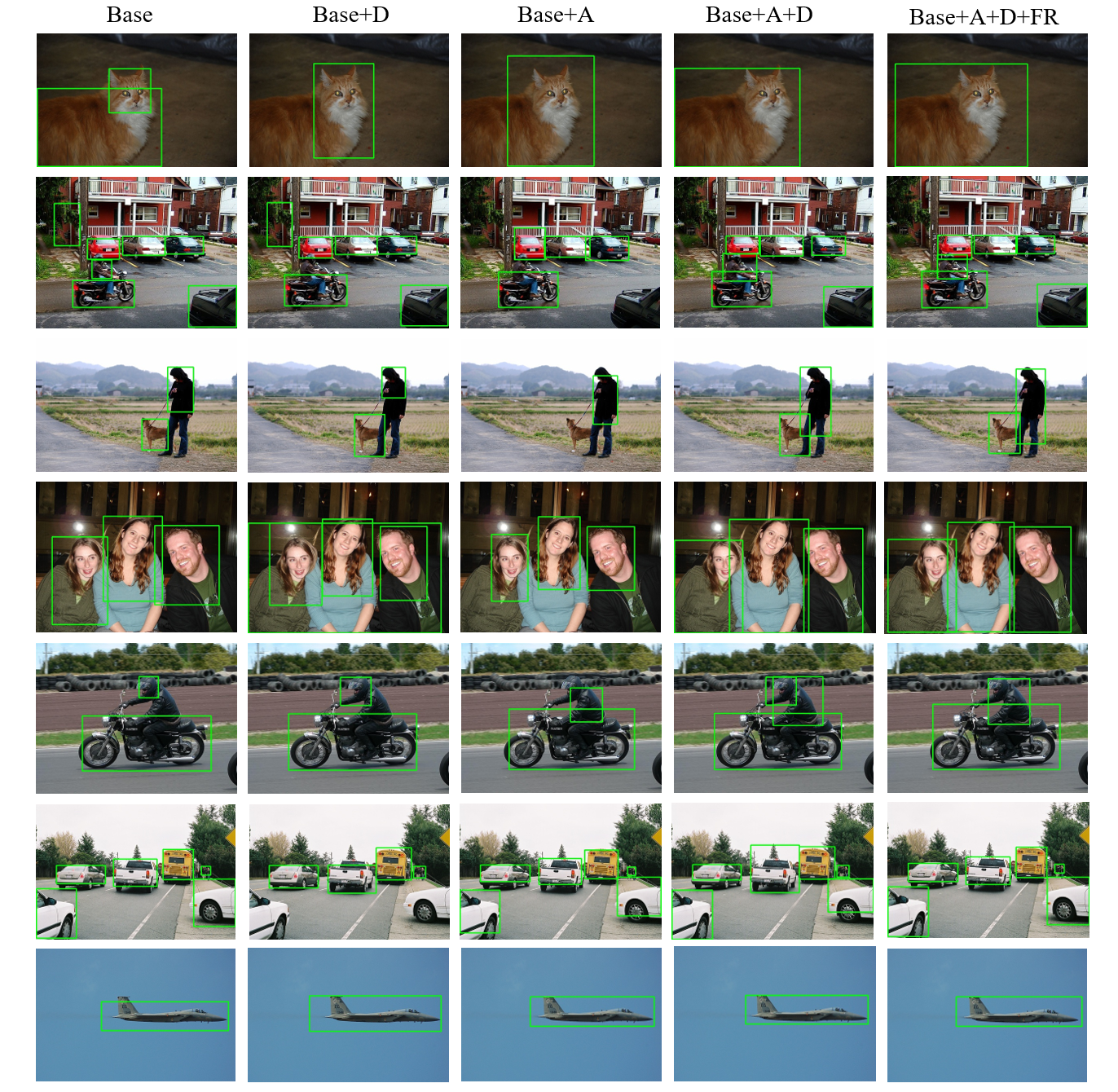}
    \caption{Visualization results of D-CPE with different components. “Base” represents the case only with the LSTM encoder, “D” represents the case that additional decoder is added, “A” represents the case that the attention mechanism is added to the LSTM encoder, “FR” represents the usage of pseudo-labels for box regression.}
    \label{fig_comp1}
\end{figure*} 

\begin{figure}[t]
    \centering
    \includegraphics[width=0.50\textwidth]{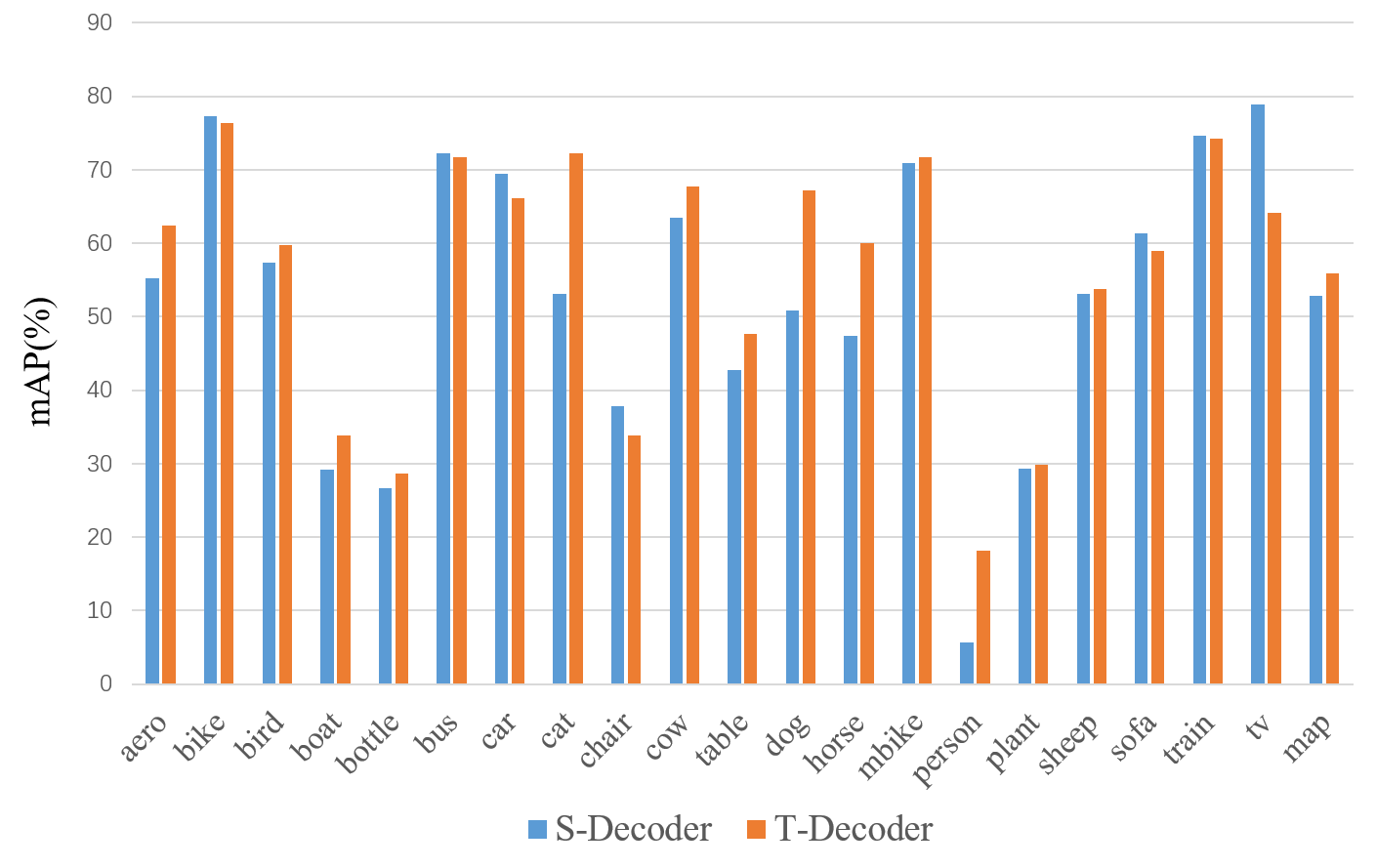}
    \caption{ The performance of “S-Decoder” and “T-Decoder” on PASCAL VOC 2007 dataset.}
    \label{DD-CPE}
\end{figure}

\begin{figure}[t]
    \includegraphics[width=0.4\textwidth]{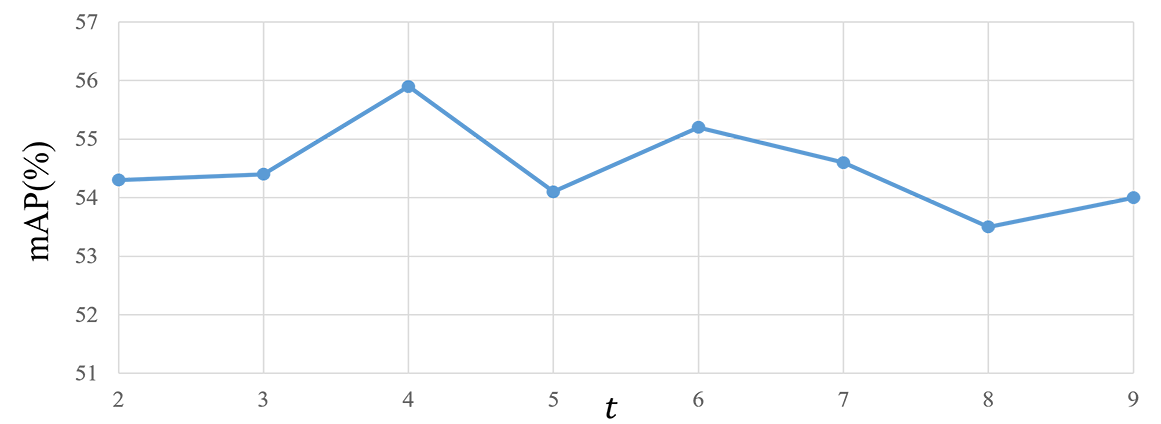}\caption{\label{infule_m}mAP (\%) with different hyper-parameters $t$ on PASCAL VOC 2007 dataset.}
\end{figure}
\begin{figure}[!t]
    \includegraphics[width=0.4\textwidth]{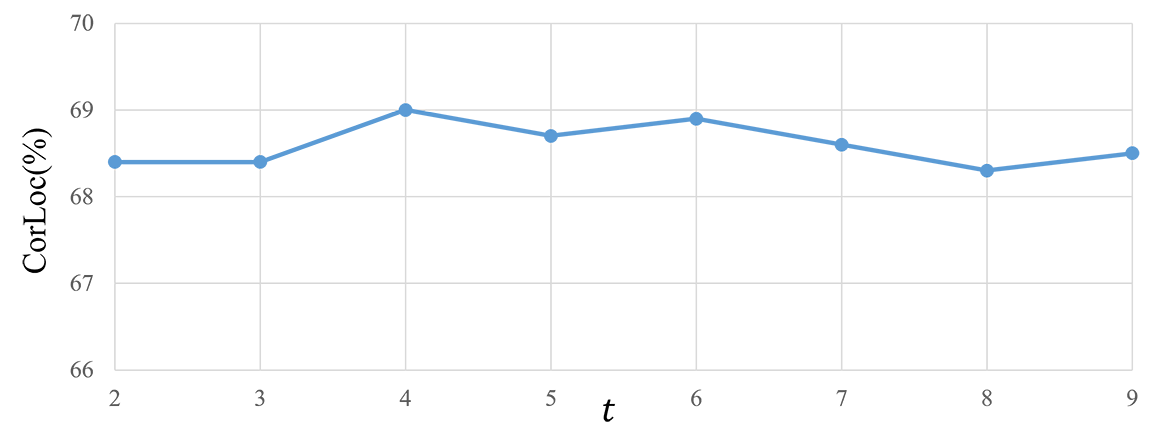}\caption{\label{infule_c}CorLoc (\%) with different hyper-parameters $t$ on PASCAL VOC 2007 dataset.}
\end{figure}

{\bf Different components of D-CPE.}
In this section, we report the influence of different D-CPE components. As shown in Tab. \ref{ablation_A}, rows 1, 2, 3 respectively show the results with “Base”, “Base+A” and “Base+D”. Row 4 shows the results of “Base+A+D”. By comparing the results in row 1 and row 2, the mAP of “Base+A” is worse than that of “Base” because the additional attention mechanism allows the LSTM encoder to select more critical information, but it may also ignore some objects in the current scene. By increasing the decoder based on dual-stream structure network to constrain the coding space of LSTM encoder, better results can be obtained than that of “Base”. The reason is that the decoder effectively restrains the encoding space, so that it can choose as many objects as possible in the scene. When “Base+A+D” is used in our method, we have obtained the best result due to the combination of LSTM network with self-attention mechanism and decoder, because the attention mechanism will capture the key information in the scene as much as possible, and the decoder requires the encoder to capture more complete scene information to include all objects in the scene as much as possible. To more intuitively observe different results generated by these components, we show some sample results in Fig. \ref{fig_comp1}. In the first four columns, we can see that in “Base” cases, the model may treat a single object as two objects (row 1) or the detector regards the background as one object (row 2). After adding decoders on “Base” to constrain the LSTM encoder, although the case where different positions of the object are contained by multiple boxes can be reduced, there may be a phenomenon that the detector generates one box for multiple similar objects (row 4) and can not detect all objects in the scene (row 6). Similar to the case of “Base+D”, in the case of “Base+A”, some objects in the scene can not be detected in row 2, 3, but get tighter boxes in row 5. In the case of “Base+A+D”, the situations mentioned above can be alleviated, and more reasonable boxes can be obtained. However, there are still some boxes that do not match the actual object or multiple boxes that are generated for a single object. The last column shows the results generated by our model trained with Fast R-CNN. 

\begin{table*}[t]
\renewcommand\arraystretch{1.2}
\caption{Results with different ${\alpha}$ on PASCAL VOC 2007 dataset} 
\scalebox{0.82}{
    \begin{tabular}{l|cccccccccccccccccccc|c}
    \hline
        \textbf{${\alpha}$} & \textbf{aero} & \textbf{bike} & \textbf{bird} & \textbf{boat} & \textbf{bottle} & \textbf{bus} & \textbf{car} & \textbf{cat} & \textbf{chair} & \textbf{cow} & \textbf{table} & \textbf{dog} & \textbf{horse} & \textbf{mbike} & \textbf{person} & \textbf{plant} & \textbf{sheep} & \textbf{sofa} & \textbf{train} & \textbf{tv} & \textbf{mAP} \\
    \hline
    \hline
       0.1 	&66.4	&78.4	&59.3	&34.2	&34.1	&74.1	&70.2	&74.2	&31.5	&63.4	&43.8	&60.7	&62.2	&70.1	&14.4	&\textbf{33.4}	&46.4	&62.0	&73.8	&62.2	&55.7\\
       0.2 &65.2	&76.4	&58.8	&35.9	&35.2	&73.0	 &69.5	&\textbf{75.8}	&29.4	&62.6	&46.6	&63.3	&63.9	&70.1	&13.9	&31.5	&42.8	&60.1	&73.1	&68.4	&55.8\\
       0.3 &63.9	&78.0	&58.2	&33.2	&33.4	&73.7	&\textbf{71.0}	&74.4	&29.9	&62.6	&45.1	&62.3	&62.3	&69.8	&16.3	&32.0	&47.2	&\textbf{62.9}	&\textbf{75.2}	&58.2	&55.5\\
       0.4 	&61.5	&78.3	&57.1	&\textbf{36.1}	&\textbf{36.4}	&74.5	&65.0	&74.6	&28.5	&61.3	&46.8	&58.7	&60.5	&70.0	&15.9	&32.3	&43.1	&59.5	&74.7	&\textbf{68.8}	&55.2\\
       0.5 & 62.4  & 76.4  & \textbf{59.7}  & 33.8  & 28.7  & 71.7  & 66.1    & 72.2  & \textbf{33.9}  & \textbf{67.7}  & 47.6  & \textbf{67.2} & 60.0 & \textbf{71.7} & \textbf{18.1} & 29.9  &  \textbf{53.8}  & 58.9    & 74.3  & 64.1    & \textbf{55.9}\\
       0.6 	&\textbf{67.0}	&76.8	&56.9	&34.5	&31.3	&73.7	&69.4	&70.9	&29.4	&61.2	&47.7	&64.5	&\textbf{65.1}	&70.7	&13.6	&32.1	&42.8	&58.8	&72.6	&58.1	&54.9\\
       0.7 &66.5	&76.7	&55.6	&32.5	&34.5	&\textbf{75.3}	&68.3	&72.4	&28.1	&61.6	&47.3	&67.0	&61.5	&69.5	&17.4	&31.9	&44.3	&58.9	&73.8	&58.2	&55.1\\
       0.8 &65.1	&77.8	&56.7	&33.1	&33.4	&74.3	&67.5	&73.9	&30.9	&64.9	&\textbf{48.6}	&66.0	&63.0	&69.9	&16.1	&30.8	&41.7	&61.0	&71.6	&60.7	&55.3\\
       0.9 	&63.3	&\textbf{78.8}	&56.5	&33.3	&33.4	&73.8	&66.9	&71.1	&29.5	&65.5	&48.2	&65.0	&60.5	&70.4	&15.2	&33.0	&41.8	&58.3	&74.6	&60.5	&55.0\\
    \hline
    \end{tabular}%
     }
     \newline
   \label{gamma_h}
\end{table*}

\begin{figure*}[t]
    \centering
    \includegraphics[width=0.9\textwidth]{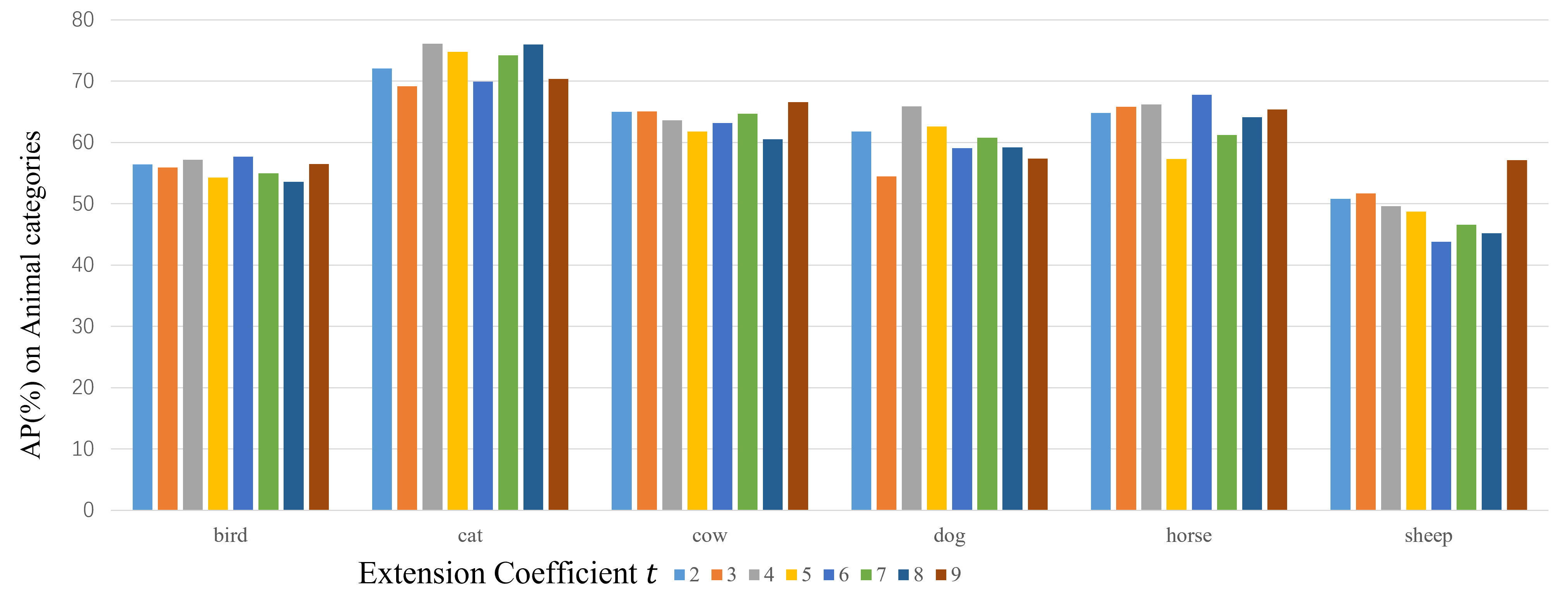}
    \captionsetup{justification=centering}
    \caption{The comparison results of the AP(\%) on animal categories with different coefficients $t$. }
    \label{Da-CPE}
\end{figure*} 
{\bf Different decoders of D-CPE.}
Fig. \ref{DD-CPE} shows that the effect of different decoders in D-CPEs. “S-Decoder” represents the decoder of the linear layer using ${sigmoid}$ as the activation function. “T-Decoder” represents the dual-stream branch network is used with $softmax$ as the activation function.

Compared to the case which constrains the LSTM encoder using “S-Decoder”, the usage of “T-Decoder” can obtain better results in more cases. In individual categories, “T-Decoder” has also shown great advantages, especially in $cat$, $dog$ and $person$ categories, the result with “T-Decoder” is at least 13 percent higher than that with “S-Decoder”. And in $aeroplane$, $bird$, $boat$, $bottle$, $cow$ and $table$ categories, it is about 3 to 8 percent higher. The results by these two decoders are similar in $bike$, $bus$, $plant$, $sheep$, and $train$ categories, with the difference of less than 3 percent. In the $car$, $chair$ and $sofa$ categories,  the “S-Decoder” performs better, especially on $tv$, which is higher than the “T-Decoder” by 8 percent. The performance of these two decoders is quite different in the $cat$, $dog$, $person$, and $tv$ categories. The reason is that $cat$, $dog$, and $person$ often appear together with other objects, while the $tv$ category often appears alone in the scene. In different scenarios, these two decoders have different performances, especially when the scene contains multiple detection objects. The “T-Decoder” can more widely select the instances that contain the object, therefore, it can add more comprehensive constraints on the encoder. In the monotonous object scenario, “S-Decoder” can effectively reduce the noise in the encoder to generate more reasonable contrastive semantics of initial proposals and extended proposals.

{\bf Coupling coefficient of D-CPE.}
Tab. \ref{gamma_h} shows the influence of different coefficient ${\alpha}$ on the different categories. On the whole, mAP is at a peak when ${\alpha}$ = 0.5, and at the lowest point when ${\alpha}$ = 0.6. With different ${\alpha}$, $aeroplane$, $bottle$, $car$, $cow$, $dog$, $horse$, $sheep$ and $tv$ categories will have large changes, which can reach up to 5 percent. The biggest difference is $tv$, when ${\alpha}$ = 0.4, AP = 68.8\%, while ${\alpha}$ = 0.6, AP = 58.1\%. When the ${\alpha}$ changes, the APs with smaller changes are $bike$ and $motorbike$, the AP gap between their maximum and minimum is only 2.4\% and 2.2\%. In most categories, the change of AP in the adjacent interval is relatively small with the increase of ${\alpha}$. This proves that different categories are sensitive to the hyper-parameter ${\alpha}$, which is caused by the aspect ratios of different categories in the scene.

\begin{table*}[htbp]\scriptsize
~~~~~~~~~
 \caption{Comparison on PASCAL VOC 2007 trainval set measured in terms of AP (\%) metric. }
\vspace{0.5em}
\scalebox{0.76}{
\renewcommand\arraystretch{1.2}
    \begin{tabular}{l|cccccccccccccccccccc|c}
    \hline
        \textbf{method} & \textbf{aero} & \textbf{bike} & \textbf{bird} & \textbf{boat} & \textbf{bottle} & \textbf{bus} & \textbf{car} & \textbf{cat} & \textbf{chair} & \textbf{cow} & \textbf{table} & \textbf{dog} & \textbf{horse} & \textbf{mbike} & \textbf{person} & \textbf{plant} & \textbf{sheep} & \textbf{sofa} & \textbf{train} & \textbf{tv} & \textbf{mAP} \\
    \hline
    \hline
    WSDDN \cite{bilen2015weakly} & 39.4  & 50.1  & 31.5  & 16.3  & 12.6  & 64.5  & 42.8  & 42.6  & 10.1  & 35.7  & 24.9  & 38.2  & 34.4  & 55.6  & 9.4   & 14.7  & 30.2  & 40.7  & 54.7  & 46.9  & 34.8 \\
    OICR \cite{tang2017multiple}  & 58.0    & 62.4  & 31.1  & 19.4  & 13.0    & 65.1  & 62.2  & 28.4  & 24.8  & 44.7  & 30.6  & 25.3  & 37.8  & 65.5  & 15.7  & 24.1  & 41.7  & 46.9  & 64.3  & 62.6  & 41.2 \\
    PCL \cite{tang2018pcl}   & 54.4  & 69.0   & 39.3  & 19.2  & 15.7  & 62.9  & 64.4  & 30.0    & 25.1  & 52.5  & 44.4  & 19.6  & 39.3  & 67.7  & 17.8  & 22.9  & 46.6  & 57.5  & 58.6  & 63.0    & 43.5 \\
    W2F \cite{2018W2F}&63.5 &70.1 &50.5 &31.9 &14.4 &72.0 &67.8 &\textbf{73.7} &23.3 &53.4 &49.4 &65.9 &57.2 &67.2 &27.6 &23.8 &51.8 &58.7 &64.0 &62.3 &52.4\\
    WS-JDS \cite{shen2019cyclic} &52.0 &64.5 &45.5 &26.7 &27.9 &60.5 &47.8 &59.7 &13.0 &50.4 &46.4 &56.3 &49.6 &60.7 &25.4 &28.2 &50.0 &51.4 &66.5 &29.7 &45.6\\
    CAP-SRN \cite{0Object} &61.5 &64.8 &43.7 &26.4 &17.1 &67.4 &62.4 &67.8 &25.4 &51.0 &33.7 &47.6 &51.2 &65.2 &19.3 &24.4 &44.6 &54.1 &65.6 &59.5 &47.6\\
    C-MIL \cite{wan2019c} & 62.5  & 58.4  & 49.5  & 32.1  & 19.8  & 70.5  & 66.1  & 63.4  & 20.0  & 60.5  & \textbf{52.9}  & 53.5  & 57.4  & 68.9  & 8.4   & 24.6  & 51.8  & 58.7  & 66.7  & 63.5  & 50.5 \\
     C-MIDN \cite{gao2019c} & 53.3  & 71.5  & 49.8  & 26.1  & 20.3  & 70.3  & 69.9  & 68.3  & 28.7  & 65.3  & 45.1  & 64.6  & 58.0    & 71.2  & \textbf{20.0}    & 27.5  & \textbf{54.9}  & 54.9  & 69.4  & 63.5  & 52.6 \\
    Pred Net \cite{arun2019dissimilarity} & \textbf{66.7}  & 69.5  & 52.8  & 31.4  & 24.7  & \textbf{74.5}  & \textbf{74.1}  & 67.3  & 14.6  & 53.0  & 46.1  & 52.9  & \textbf{69.9}  & 70.8  & 18.5  & 28.4  & 54.6  & \textbf{60.7}  & 67.1  & 60.4  & 52.9 \\
    SDCN \cite{0Weakly} &59.4 &71.5 &38.9 &32.2 &21.5 &67.7 &64.5 &68.9 &20.4 &49.2 &47.6 &60.9 &55.9 &67.4 &31.2 &22.9 &45.0 &53.2 &60.9 &64.4 &50.2\\
    OIM+IR \cite{2020Object} &55.6 &67.0 &45.8 &27.9 &21.1 &69.0 &68.3 &70.5 &21.3 &60.2 &40.3 &54.5 &56.5 &70.1 &12.5 &25.0 &52.9 &55.2 &65.0 &63.7 &50.1\\
    SLV \cite{chen2020slv}   & 65.6  & 71.4  & 49.0  & \textbf{37.1}  & 24.6  & 69.6  & 70.3  & 70.6  & 30.8  & 63.1  & 36.0  & 61.4  & 65.3  & 68.4  & 12.4  &  \textbf{29.9}  & 52.4  & 60.0  & 67.6  & 64.5  & 53.5 \\
    PG-PS \cite{cheng2020high} & 63.0  & 64.4  & 50.1  & 27.5  & 17.1  & 70.6  & 66.0  & 71.1  & 25.8  & 55.9  & 43.2  & 62.7  & 65.9  & 64.1  & 10.2  & 22.5  & 48.1  & 53.8  & 72.2  & \textbf{67.4}  & 51.1 \\
    CASD\cite{Huang2020ComprehensiveAS} & -  & -  & -  & -  & -  & -  & -  & -  & -  & -  & -  & -  & -    & -  & -    & -  & -  & -  & -  & -  & 55.3 \\
    P-MIDN+MGSC \cite{2021Pyramidal} & -  & -  & -  & -  & -  & -  & -  & -  & -  & -  & -  & -  & -    & -  & -    & -  & -  & -  & -  & -  & 53.9 \\
    
    \hline
    CPE(Ours) & 62.4  & \textbf{76.4}  & \textbf{59.7}  & 33.8  & \textbf{28.7}  & 71.7  & 66.1    & 72.2  & \textbf{33.9}  & \textbf{67.7}  & 47.6  & \textbf{67.2} & 60.0 & \textbf{71.7} & 18.1 & \textbf{29.9}  & 53.8  & 58.9    & \textbf{74.3}  & 64.1    & \textbf{55.9} \\
    \hline
    OICR+FRCNN \cite{tang2017multiple} & 65.5  & 67.2  & 47.2  & 21.6  & 22.1  & 68.0    & 68.5  & 35.9  & 5.7   & 63.1  & 49.5  & 30.3  & 64.7  & 66.1  & 13.0    & 25.6  & 50.0    & 57.1  & 60.2  & 59.0    & 47.0 \\
    PCL+FRCNN \cite{tang2018pcl}& 63.2  & 69.9  & 47.9  & 22.6  & 27.3  & 71.0    & 69.1  & 49.6  & 12.0    & 60.1  & 51.5  & 37.3  & 63.3  & 63.9  & 15.8  & 23.6  & 48.8  & 55.3  & 61.2  & 62.1  & 48.8 \\
    WS-JDS+FRCNN\cite{shen2019cyclic}&64.8 &70.7 &51.5 &25.1 &29.0 &74.1 &69.7 &69.6 &12.7 &\textbf{69.5} &43.9 &54.9 &39.3 &71.3 &32.6 &29.8 &\textbf{57.0} &61.0 &66.6 &57.4 &52.5\\
    CAP-SRN+FRCNN \cite{0Object} & - & - & - & - & - & - & - & - & - & - & - & - & - & - & - & - & - & - & - & - & 52.1\\
    C-MIL+FRCNN \cite{wan2019c}   & 61.8  & 60.9  & 56.2  & 28.9  & 18.9  & 68.2  & 69.6  & 71.4  & 18.5  & 64.3  & \textbf{57.2}  & 66.9  & 65.9  & 65.7  & 13.8  & 22.9  & 54.1  & 61.9  & 68.2  & 66.1  & 53.1 \\
    Pred Net(ENS) \cite{arun2019dissimilarity} & \textbf{67.7}  & 70.4  & 52.9  & 31.3  & 26.1  & \textbf{75.5}  & \textbf{73.7}  & 68.6  & 14.9  & 54.0  & 47.3  & 53.7  & 70.8  & 70.2  & 19.7  & 29.2  & 54.9  & 61.3  & 67.6  & 61.2  & 53.6 \\
    SDCN+FRCNN \cite{0Weakly} &59.8 &75.1 &43.3 &31.7 &22.8 &69.1 &71.0 &72.9 &21.0 &61.1 &53.9 &73.1 &54.1 &68.3 &\textbf{37.6} &20.1 &48.2 &62.3 &67.2 &61.1 &53.7\\
    OIM+IR+FRCNN \cite{2020Object} &53.4 &72.0 &51.4 &26.0 &27.7 &69.8 &69.7 &74.8 &21.4 &67.1 &45.7 &63.7 &63.7 &67.4 &10.9 &25.3 &53.5 &60.4 &70.8 &58.1 &52.6\\
     C-MIDN+FRCNN\cite{gao2019c} & 54.1  & 74.5  & 56.9  & 26.4  & 22.2  & 68.7  & 68.9  & 74.8  & 25.2  & 64.8  & 46.4  & 70.3  & 66.3  & 67.5  & 21.6  & 24.4  & 53.0    & 59.7  & 68.7  & 58.9  & 53.6 \\
    SLV+FRCNN \cite{chen2020slv} & 62.1  & 72.1  & 54.1  & 34.5  & 25.6  & 66.7  & 67.4  & \textbf{77.2}  & 24.2  & 61.6  & 47.5  & 71.6  & \textbf{72.0}  & 67.2  & 12.1  & 24.6  & 51.7  & 61.1  & 65.3  & 60.1  & 53.9 \\
    PG-PS+FRCNN \cite{cheng2020high} & 59.3  & 66.2  & 55.4  & \textbf{35.2}  & 22.3  & 69.7  & 70.2  & 73.8  & 29.4  & 63.6  & 47.9  & \textbf{78.1}  & 67.9  & 68.2  & 12.2  & 24.9  & 43.2  & \textbf{63.7}  & 73.2  & \textbf{66.8}  & 54.6 \\  
    MiST(+reg) \cite{ren2020instance}& - & - & - & - & - & - & - & - & - & - & - & - & - & - & - & - & - & - & - & - & 54.9 \\    
    CASD\cite{Huang2020ComprehensiveAS} & -  & -  & -  & -  & -  & -  & -  & -  & -  & -  & -  & -  & -    & -  & -    & -  & -  & -  & -  & -  & 56.8 \\
    
    P-MIDN+MGSC+FRCNN \cite{2021Pyramidal} & -  & -  & -  & -  & -  & -  & -  & -  & -  & -  & -  & -  & -    & -  & -    & -  & -  & -  & -  & -  & 55.0 \\
    \hline
    CPE+FRCNN(Ours) & 66.0  & \textbf{78.7}  & \textbf{59.7}  & 34.0  & \textbf{33.1}  & 73.7  & 69.6  & 75.1  & \textbf{33.1}  & 64.4  & 50.3  & 69.1    & 68.2  & \textbf{70.4}  &19.6  & \textbf{31.6}  & 49.9 & 62.4  & \textbf{74.5}  & 62.3    & \textbf{57.3} \\
    \hline
    \end{tabular}%
}
\label{voc07map}
\end{table*}%
\begin{table*}[t]\scriptsize

\caption{Comparison on PASCAL VOC 2007 trainval set measured in terms of CorLoc (\%) metric. }
\scalebox{0.75}{
\renewcommand\arraystretch{1.2}
    \begin{tabular}{l|cccccccccccccccccccc|c}
    \hline
    \textbf{method} & \textbf{aero} & \textbf{bike} & \textbf{bird} & \textbf{boat} & \textbf{bottle} & \textbf{bus} & \textbf{car} & \textbf{cat} & \textbf{chair} & \textbf{cow} & \textbf{table} & \textbf{dog} & \textbf{horse} & \textbf{mbike} & \textbf{person} & \textbf{plant} & \textbf{sheep} & \textbf{sofa} & \textbf{train} & \textbf{tv} & \textbf{CorLoc} \\
    \hline
    \hline
    WSDDN \cite{bilen2015weakly} & 65.1  & 58.8  & 58.5  & 33.1  & 39.8  & 68.3  & 60.2  & 59.6  & 34.8  & 64.5  & 30.5  & 43    & 56.8  & 82.4  & 25.5  & 41.6  & 61.5  & 55.9  & 65.9  & 63.7  & 53.5 \\
    OICR \cite{tang2017multiple}  & 81.7  & 80.4  & 48.7  & 49.5  & 32.8  & 81.7  & 85.4  & 40.1  & 40.6  & 79.5  & 35.7  & 33.7  & 60.5  & 88.8  & 21.8  & 57.9  & 76.3  & 59.9  & 75.3  & 81.4  & 60.6 \\
    PCL(baseline) \cite{tang2018pcl}   & 79.6  & 85.5  & 62.2  & 47.9  & 37.0    & 83.8  & 83.4  & 43.0    & 38.3  & 80.1  & 50.6  & 30.9  & 57.8  & 90.8  & 27.0    & 58.2  & 75.3  & \textbf{68.5}  & 75.7  & 78.9  & 62.7 \\
    WS-JDS \cite{shen2019cyclic}  &82.9 &74.0 & \textbf{73.4} &47.1 &\textbf{60.9} &80.4 &77.5 &78.8 &18.6 &70.0 &56.7 &67.0 &64.5 &84.0 &47.0 &50.1 &71.9 &57.6 &83.3 &43.5 &64.5\\
    CAP-SRN \cite{0Object} &85.5 &79.6 &68.1 &55.1 &33.6 &83.5 &83.1 &78.5 &42.7 &79.8 &37.8 &61.5 &74.4 &88.6 &32.6 &55.7 &77.9 &63.7 &78.4 &74.1 &66.7\\
    Pred Net \cite{arun2019dissimilarity} & \textbf{88.6}  & \textbf{86.3}  & 71.8  & 53.4  & 51.2  & \textbf{87.6}  & \textbf{89.0}    & 65.3  & 33.2  & \textbf{86.6}  & \textbf{58.8}  & 65.9  & 87.7  & \textbf{93.3}  & 30.9  & 58.9  & 83.4  & 67.8  & 78.7  & 80.2  & 70.9 \\
    C-MIDN \cite{gao2019c}& - & - & - & - & - & - & - & - & - & - & - & - & - & - & - & - & - & - & - & -     & 68.7 \\
    SDCN \cite{0Weakly} &85.0 &83.9 &58.9 & \textbf{59.6} &43.1 &79.7 &85.2 &77.9 &31.3 &78.1 &50.6 &75.6 &76.2 &88.4 & \textbf{49.7} &56.4 &73.2 &62.6 &77.2 &79.9 &68.6\\
    OIM+IR \cite{2020Object} & - & - & - & - & - & - & - & - & - & - & - & - & - & - & - & - & - & - & - & -     & 67.2 \\
    SLV \cite{chen2020slv}  & 84.6  & 84.3  & 73.3  & 58.5  & 49.2  & 80.2  & 87.0    & 79.4  & 46.8  & 83.6  & 41.8  & 79.3  & \textbf{88.8}  & 90.4  & 19.5  & 59.7  &  \textbf{79.4}  & 67.7  & 82.9  & \textbf{83.2}  & \textbf{71.0}  \\
    PG-PS \cite{cheng2020high} & 85.4  & 80.4  & 69.1  & 58.0  & 35.9  & 82.7  & 86.7  &\textbf{ 82.6}  & 45.5  & 84.9  & 44.1  & \textbf{80.2}  & 84.0  & 89.2  & 12.3  & 55.7  & \textbf{79.4}  & 63.4  & 82.1  & 82.1  & 69.2 \\
    P-MIDN+MGSC \cite{2021Pyramidal} & -  & -  & -  & -  & -  & -  & -  & -  & -  & -  & -  & -  & -    & -  & -    & -  & -  & -  & -  & -  & 69.8 \\
    \hline
    CPE(Ours) & 82.9  & 82.4  & 70.3  & 53.7  & 43.5  & 81.7  & 80.2  & 77.0  & \textbf{51.7}  & 82.9    & 46.8  & 75.1  & 74.5  & 89.6 & 29.4  & \textbf{60.8}  & 77.3  & 60.2  & \textbf{83.3}  & 76.7  & 69.0 \\
    \hline
    OICR+FRCNN \cite{tang2017multiple} & 85.8  & 82.7  & 62.8  & 45.2  & 43.5  & 84.8  & 87    & 46.8  & 15.7  & 82.2  & 51.0    & 45.6  & 83.7  & 91.2  & 22.2  & 59.7  & 75.3  & 65.1  & 76.8  & 78.1  & 64.3 \\
    PCL+FRCNN(baseline) \cite{tang2018pcl} & 83.8  & 85.1  & 65.5  & 43.1  & 50.8  & 83.2  & 85.3  & 59.3  & 28.5  & 82.2  & 57.4  & 50.7  & 85.0  & 92.0  & 27.9  & 54.2  & 72.2  & 65.9  & 77.6  & 82.1  & 66.6 \\
    WS-JDS+FRCNN\cite{shen2019cyclic} &79.8 &84.0 &68.3 &40.2 & \textbf{61.5} &80.5 &85.8 &75.8 &29.7 &77.7 &49.5 &67.4 &58.6 &87.4 &66.2 &46.6 &78.5 & \textbf{73.7} &84.5 &72.8 &68.6\\
    CAP-SRN+FRCNN \cite{0Object} & - & - & - & - & - & - & - & - & - & - & - & - & - & - & - & - & - & - & - & -     & - \\
    Pred Net(ENS) \cite{arun2019dissimilarity}& \textbf{89.2}  & \textbf{86.7}  & 72.2  & 50.9  & 51.8  & \textbf{88.3}  & \textbf{89.5}  & 65.6  & 33.6  & \textbf{87.4}  & \textbf{59.7}  & 66.4  & 88.5  & \textbf{94.6}  & 30.4  & 60.2  &  \textbf{83.8}  & 68.9  & 78.9  & 81.3  & 71.4 \\
    C-MIDN+FRCNN \cite{gao2019c}& - & - & - & - & - & - & - & - & - & - & - & - & - & - & - & - & - & - & - & -   & 71.9 \\
    SDCN+FRCNN \cite{0Weakly} &85.0 &86.7 &60.7 &62.8 &46.6 &83.2 &87.8 &81.7 &35.8 &80.8 &57.4 &81.6 &79.9 &92.4 &59.3 &57.5 &79.4 &68.5 &81.7 &81.4 &\textbf{72.5}\\
    OIM+IR+FRCNN \cite{2020Object} & - & - & - & - & - & - & - & - & - & - & - & - & - & - & - & - & - & - & - & -     & 68.8 \\
    
    SLV+FRCNN \cite{chen2020slv}& 85.8  & 85.9  & 73.3  & 56.9  & 52.7  & 79.7  & 87.1  & 84.0  & 49.3  & 82.9  & 46.8  & 81.2  & \textbf{89.8}  & 92.4  & 21.2  & 59.3  & 80.4  & 70.4  & 82.1  & 78.8  & 72.0  \\
    PG-PS+FRCNN \cite{cheng2020high}& 87.1  & 84.4  & 70.6  & \textbf{57.7}  & 46.1  & 85.7  & 88.1  & \textbf{85.6}  & 46.7  & 87.2  & 45.9  & \textbf{83.4}  & 85.6  & 90.1  & 18.1  & 59.7  & 82.4  & 68.2  & \textbf{85.3}  & \textbf{86.1}  & 72.2 \\
    
    P-MIDN+MGSC+FRCNN \cite{2021Pyramidal} & -  & -  & -  & -  & -  & -  & -  & -  & -  & -  & -  & -  & -    & -  & -    & -  & -  & -  & -  & -  & 72.4 \\

    \hline
    CPE+FRCNN(Ours) & 82.1  & 86.3  & \textbf{76.3}  & 55.9  &48.9  & 84.8  & 84.1  & 79.4  & \textbf{51.2}  & 84.2  & 52.5  & 77.9 &80.3  & 93.2  & \textbf{32.0}  & \textbf{64.1}  & 82.5  & 65.9  & 85.2  & 80.6  & 72.4 \\
    \hline
    \end{tabular}%
  }
      \newline
\label{voc07cor}
\end{table*}%

\begin{table}[h]
\caption{Detection and localization performance (\%) on PASCAL VOC 2012 dataset.}
\centering
    \begin{tabular}{l|c|c c}
    \hline
         \textbf{method} & \textbf{mAP(\%)} & \textbf{CorLoc(\%)}\\
    \hline
    \hline
    OICR \cite{tang2017multiple}  & 37.9 & 62.1\\
    PCL \cite{tang2018pcl}  & 40.6 & 63.2\\
    C-MIL \cite{wan2019c}  & 46.7 & 67.4\\
    WS-JDS \cite{shen2019cyclic}  & 39.1 & 63.5\\
    C-MIDN  \cite{gao2019c}    & 50.2 & 71.2\\
    SDCN \cite{0Weakly} &43.5 &67.9\\
    PG-PS \cite{cheng2020high}  & 48.3 & 68.7\\
    OIM+IR \cite{2020Object}    & 45.3 & 67.1\\
    MIST  \cite{ren2020instance}    &52.1  & 70.9\\
    P-MIDN+MGSC \cite{2021Pyramidal}   & 52.8 & 73.3\\
    \hline
    CPE  &\textbf{54.3\footnotemark[1]} &\textbf{73.5}\\
    \hline
    OICR+FRCNN \cite{tang2017multiple}  & 42.5 & 65.6\\
    PCL+FRCNN(baseline) \cite{tang2018pcl}   & 40.6 & 63.2\\
    Pred Net \cite{arun2019dissimilarity}  & 48.4 & 69.5\\
    SDCN+FRCNN \cite{0Weakly} &46.7 &69.5\\
    PG-PS+FRCNN \cite{cheng2020high}    & 52.9 & 71.0\\
    SLV+FRCNN \cite{chen2020slv}  & 49.2 & 69.2\\
    %MIST+FRCNN \cite{ren2020instance}    & 52.1 & 70.9\\
    C-MIDN+FRCNN \cite{gao2019c}   & 50.3 & 73.3\\
    CASD\cite{Huang2020ComprehensiveAS} &53.6 &-\\
    P-MIDN+MGSC+FRCNN \cite{2021Pyramidal} &53.4 &\textbf{76.7}\\
    \hline
    CPE+FRCNN     & \textbf{54.6\footnotemark[2]} & 75.5\\
    \hline
    \end{tabular}%
\label{voc12}
\end{table}%

\begin{table}[h]
\caption{Experiment results of different methods on MS COCO dataset.}
\centering
\scalebox{0.85}{
\renewcommand\arraystretch{1.2}
    \begin{tabular}{l|c c | c c c}
    \hline
         \textbf{method} & \textbf{Training data} &\textbf{Testing data} & \textbf{AP} &\textbf{AP$_{50}$} \\
    \hline
    PCL\cite{tang2018pcl} &train2014  &val2014 & 8.5 & 19.4\\
    WS-JDS \cite{2019Cyclic} &train2017  &minival & 10.5 & 20.3\\
    C-MIDN \cite{gao2019c} &train2014  &val2014 & 9.6 & 21.4\\
    PG-PS\cite{cheng2020high} &trainval135k  &minival &- &20.7\\
    P-MIDN+MGSC\cite{2021Pyramidal} &train2014  &val2014 &13.1 &\textbf{27.4}\\
    P-MIDN+MGSC\cite{2021Pyramidal} &train2014  &minival &13.4 &\textbf{27.7}\\
    \hline
    CPE  &train2014  &val2014  &\textbf{13.5} & 26.7\\
    CPE  &train2014  &minival  & \textbf{13.7} & 27.1\\
    \hline
    PCL+FRCNN \cite{tang2018pcl}   &train2014  &val2014 & 9.2 & 19.6\\
    C-MIDN+FRCNN\cite{gao2019c} &train2014  &val2014 & 9.6 & 21.4\\
    WSOD${^2}$+FRCNN\cite{zeng2019wsod2}  &train2014  &val2014 & 10.8 & 22.7\\
    MIST+FRCNN\cite{ren2020instance}  &train2014  &val2014  & 11.4 & 24.3\\
    CASD+FRCNN\cite{Huang2020ComprehensiveAS}  &train2014  &val2014 &12.8 &26.4\\
    P-MIDN+MGSC+FRCNN\cite{2021Pyramidal} &train2014  &val2014 &13.2 &\textbf{28.5}\\
    P-MIDN+MGSC+FRCNN\cite{2021Pyramidal} &train2014  &minival &13.5 &\textbf{29.0}\\
    \hline
    CPE+FRCNN  &train2014  &val2014   & \textbf{13.8} & 27.8\\
    CPE+FRCNN  &train2014  &minival  & \textbf{13.9} &28.3 \\
    \hline
    \end{tabular}%
    }
        \newline
\label{coco2014}
\end{table}%

{\bf Extension coefficient of extended proposal.}
We further investigate the effectiveness of extension coefficient. In Fig. \ref{infule_m} and \ref{infule_c}, we show the mAP and CorLoc under different extension coefficient $t$ on PASCAL VOC 2007. Obviously, we can observe when the extension coefficient $t$ increases from 2 to 4, the mAP and CorLoc become better. When $t$ is in the range of 4 to 5, CorLoc is slowly decreasing, and mAP is also decreasing. As the extension coefficient $t$ changes, the overall results of certain categories do not change much, such as $plant$ category. The best AP is 31.8\% ($t$ = 6), while the worst is 28.4\% ($t$ = 3). Some categories may show huge differences, the highest AP of $tv$ category is 69.1\% ($t$ = 6), but the worst is 43.8\% ($t$ = 1), the gap between them is about 26.0\%. The APs of other categories fluctuate continuously with the change of $t$. Different extension coefficients will cause large changes of the mAP on some categories. The reason is that if fixed coefficients are used to expand the proposal, different aspect ratios of the initial proposal will not be taken into account. Therefore, we further design an adaptive aspect ratio coefficient in Eq. \ref{5} to reduce this negative impact, resulting in an increase in mAP. As shown in Fig. \ref{Da-CPE}, we show the influence of the extension coefficient $t$ on different animal categories. With different $t$, the APs of $dog$, $horse$ and $sheep$ categories have significant changes, while the $bird$, $cat$ and $cow$ have not changed too much. The sensitivities of different categories to $t$ are due to the fact that the initial proposals of different categories have different aspect ratios in the image. 

\subsection{Comparison with the state-of-the-arts}
We compare the proposed method with other state-of-the-arts, including MIST \cite{ren2020instance}, PG-PS \cite{cheng2020high}, SLV \cite{chen2020slv} and CASD \cite{Huang2020ComprehensiveAS}, as well as other approaches utilizing segmentation for WSOD. Following previous WSOD work, our proposed method is also compared in two settings. The first one is to directly use the generated results without bounding-box regression. The second setting is to use the generated boxes by our method as pseudo ground-truths to train Fast R-CNN detector \cite{2015Fast}. Tab. \ref{voc07map} and Tab. \ref{voc07cor} show the mAP and CorLoc performance respectively on Pascal VOC 2007 dataset. The top part shows the results obtained by training our model with the first setting. Among them, our method obtains the excellent results. Compared with the mAP of baseline (OICR), our method improves by 14.7 percent. In the bottom part of the table, we show the results with the second setting. Compared to these methods trained with Fast R-CNN, our model still get the best performance $i.e.$, 57.3\% mAP among all the methods. Our method outperforms other powerful methods, MIST+FRCNN, PG-PS+FRCNN, P-MIDN+MGSC+FRCNN, by 2.4 percent, 2.7 percent and 2.3 percent in mAP respectively. In Tab. \ref{voc12}, we show the results on Pascal VOC 2012. Our method reaches mAP$_{0.5}$ of 54.6\%, which is 12.1 percent higher than that of baseline in mAP and exceeds other results of the state-of-the-arts. To further demonstrate the effectiveness of the proposed method, we conducted experiments on the more challenging MS-COCO dataset and report results in Tab. \ref{coco2014}. Our proposed method achieves AP@[.50: .05: .95] of 13.8\%, surpassing other methods by a large margin on the $val2014$ split. This shows our method can still achieve excellent results on large-scale datasets.

We also have to admit that there is a certain gap between the proposed method and other methods in Tab. \ref{voc07cor} and Tab. \ref{voc12} in the whole CorLoc. In the proposed method, as the region is gradually approaching complete, the instance may contain more general features which may belong to multiple categories. Some instances are similar to certain objects but belong to other categories. However, the proposed method recognizes them as the same class of objects, which leads to wrong judgments and increases the number of false positives (FP). At the same time, since the instance already contains regions with discriminative features, it increases the number of true positives (TP) and reduces the number of false negatives (FN). For different metrics, CorLoc only considers Precision, but the mAP considers both Precision and Recall. Therefore, our mAP is higher than that of other WSOD methods, while the CorLoc score is not the best, but still competitive.

\footnotetext[1]{\url{http://host.robots.ox.ac.uk:8080/anonymous/ZN8SAF.html}}
\footnotetext[2]{\url{http://host.robots.ox.ac.uk:8080/anonymous/K2EBNT.html}}

\subsection{Visualization Results}
\begin{figure*}[t]
    \centering
    \includegraphics[width=0.95\textwidth]{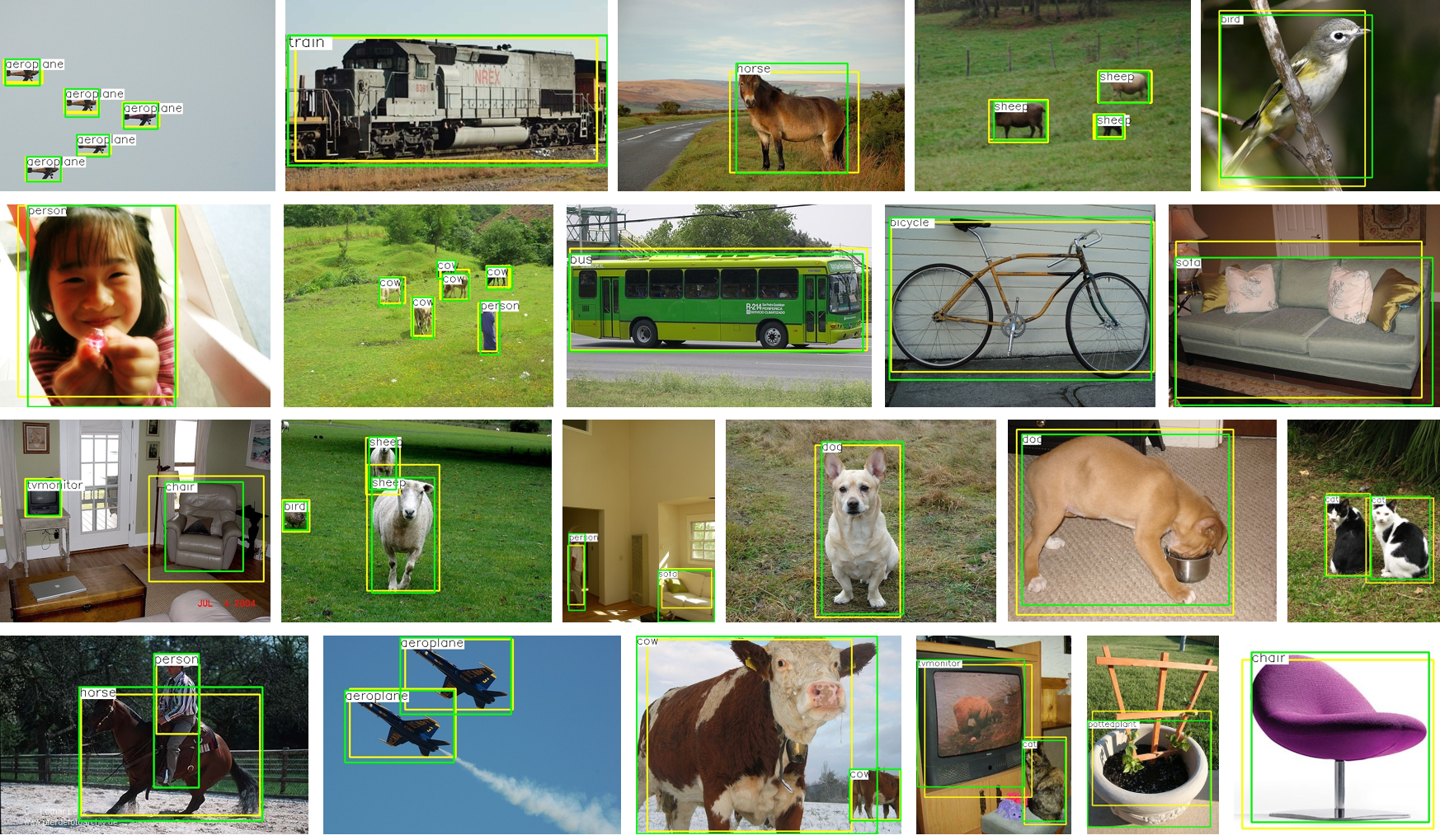}
    \caption{Successful detection results of our method. The yellow boxes are the results detected by our method, and green boxes indicate the ground-truth.}
    \label{result}
\end{figure*}

\begin{figure}[t]
    \centering
    \includegraphics[width=0.45\textwidth]{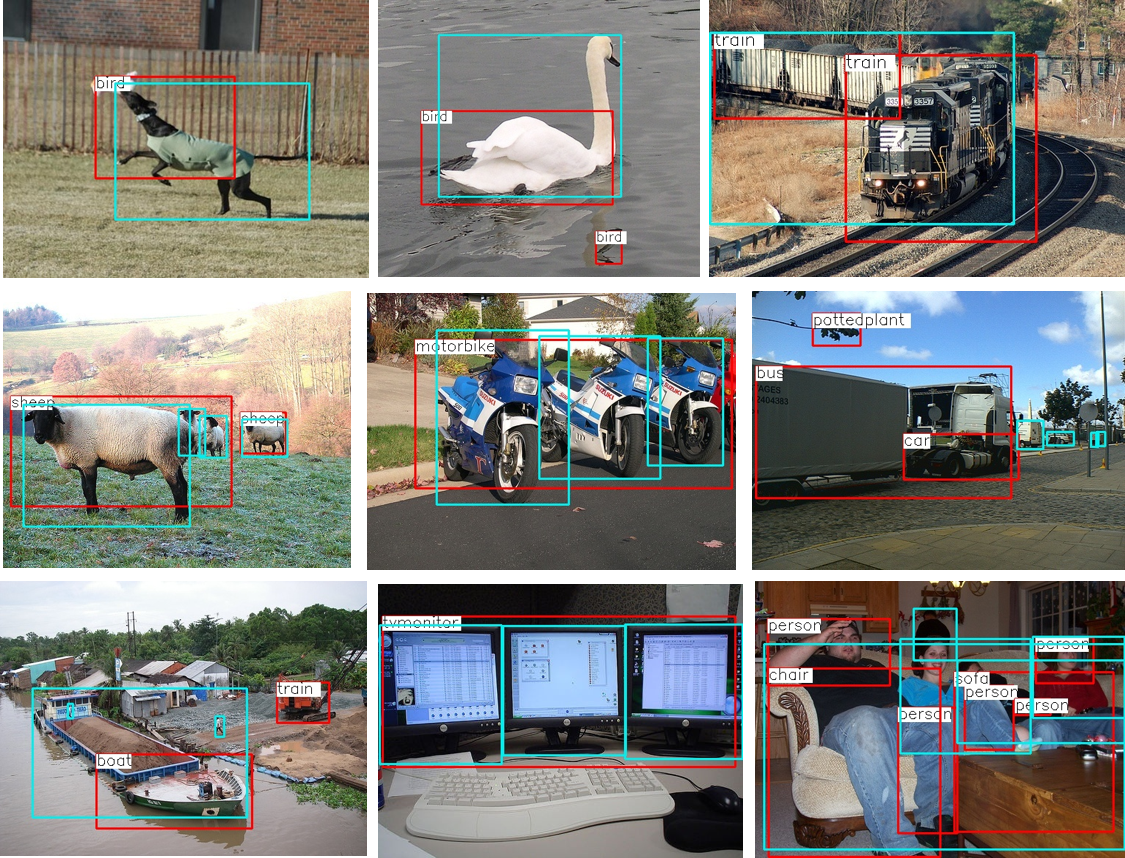}
    \caption{Failed detection results of our method. The red boxes are the results detected by our method, and blue boxes indicate the ground-truth.}
    \label{f_fig}
\end{figure}

In order to better analyze the detection results, we visualize some example images from the PASCAl VOC 2007 dataset. In Fig. \ref{result}, we show the successful detection (IOU\textgreater0.5), our method can obtain the detection results which are close to the ground-truth, and can detect the main object in most scenes. It is noted that our method can alleviate most of the mentioned problems and improve detection performance. We further show some failed detection results in Fig. \ref{f_fig}. For example, $motorbike$ and $tv$ in the second column, although the scene contains multiple similar objects, only a single detection box was generated. Similarly, in the first column, the detection box for the $sheep$ contains multiple $sheep$ objects. The reason is that the same object in the scene is too compact, and at the same time, the extension is relatively large, which makes the encoder ignore the background area of the extension. Besides, the detection boxes generated by the model are too small to contain the complete object. In the first column, the detection box for the $boat$ is too small. In addition, the detection of the $dog$ as a $bird$ in the first row shows the problem of wrong detection, and the last column in the second row shows the problem of confusing the background as the object. Compared with the CNN network, although we introduce the decoder to constrain the encoding space of the LSTM network, it still does not perform very well. And the background of images in the training data often contain some objects similar to the detection objects, which will also affect the detection accuracy and can be seen in the last column of the second row.

\section{Conclusion}

In this paper, we propose an effective WSOD method by combining the sequential encoder LSTM. We regard the initial instance and the extended one as a mutual relational network, and regard the proposal features generated by the instances as time sequences. CPE extracts the contrastive contextual semantics in different directions between the initial proposal and the extended proposal by LSTM encoders, and combines the contrastive contextual semantics with the basic WSDDN network to produce more reasonable initial instances to perform the task of WSOD. By comparing the initial proposal and the extended proposal, CPE provides a new way to optimize proposal scores. Our proposed network framework has obtained more accurate localization results, significantly improved the baseline method, and achieved the most advanced results on the PASCAL VOC 2007, 2012 and MS-COCO datasets. 

However, there are still some problems that need to be solved in the future. For example, when similar objects are enclosed together or the distance of them is too close, the CPE may recognize them as the one object, resulting in the degradation of detection performance. In addition, we will also want to explore the extension scales of different categories in different directions delicately, including more expansion methods and the combination of contrastive semantics in different directions. Furthermore, we will explore better dynamic methods to obtain extended proposals to reduce the impact of different pooling features of these proposals on extracting contrastive  semantics of the integrity.

\ifCLASSOPTIONcaptionsoff
  \newpage
\fi

\bibliographystyle{IEEEtran}
\bibliography{F_Tip_Paper_jrnl}

%\newpage
\begin{IEEEbiography}[{\includegraphics[width=1in,height=1.25in,clip,keepaspectratio]{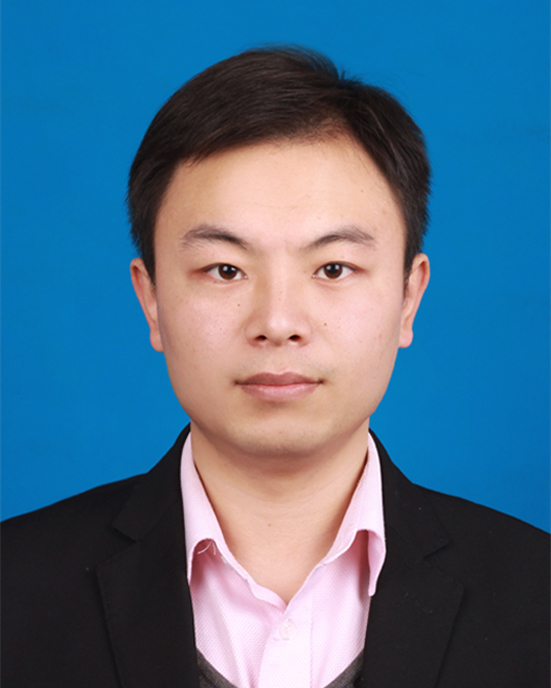}}]{Pei Lv}
received the Ph.D. degree from the State Key Laboratory of CAD\&CG, Zhejiang University Hangzhou, China, in 2013. He is an Associate Professor with the School of Computer and Artificial Intelligence, Zhengzhou University, Zhengzhou, China. His research interests include computer vision and computer graphics. He has authored more than 50 journal and conference papers in the above areas, including the IEEE Transactions on Image Processing, the IEEE Transactions on Multimedia, the IEEE Transactions on Circuits and Systems for Video Technology, CVPR, ECCV, ACM MM, and IJCAI.
\end{IEEEbiography}
\vspace{-100 mm}
\begin{IEEEbiography}[{\includegraphics[width=1in,height=1.25in,clip,keepaspectratio]{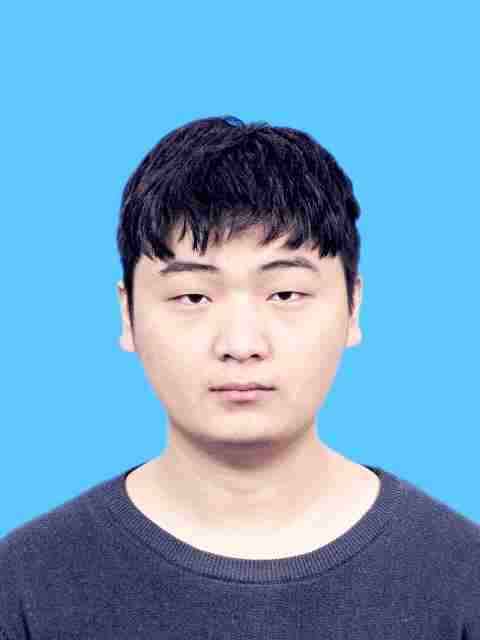}}]{Suqi Hu}
received the B.S. degree in Software Engineering, Liaoning Petrochemical University, Fushun, China. He is currently pursuing the Master degree in Henan Institute of Advanced Technology, Zhengzhou University, Zhengzhou, China. His current research interests include machine learning, computer vision and their applications to weakly supervised object detection.
\end{IEEEbiography}
\vspace{-100mm}
\begin{IEEEbiography}[{\includegraphics[width=1in,height=1.25in,clip,keepaspectratio]{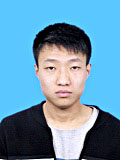}}]{Tianran Hao}
received the B.S. degrees  in School of Information Engineering, Zhengzhou University, Zhengzhou, China. He is currently pursuing the Ph.D. degree in School of Computer and Artificial Intelligence, Zhengzhou University, Zhengzhou, China. His current research interests include machine learning, computer vision and their applications for intelligent detection systems.
\end{IEEEbiography}

\end{document}